\newif\ifcl
\clfalse

\ifcl
\documentclass[shortpaper]{clv3}
\else
\documentclass{article}
\usepackage[totalwidth=460pt, totalheight=680pt]{geometry}
\usepackage{natbib}
\fi

\usepackage{adjustbox}
\usepackage[english,russian]{babel}
\usepackage{booktabs}
\usepackage{caption}
\usepackage{CJKutf8} 
\usepackage{comment}
\usepackage[T2A]{fontenc}
\usepackage{graphicx}
\usepackage{hyperref}
\usepackage[utf8]{inputenc}
\usepackage{makecell}
\usepackage{xcolor}
\usepackage{xspace}

\graphicspath{ {./figures/} }

\newcommand{\LP}{LivePerson\xspace}

\newcommand{\TODO}[1]{\noindent \textcolor{green}{TODO: #1}}
\newcommand{\rrc}{Reciprocal Rank Classifier\xspace}
\newcommand{\fasttext}{\texttt{fastText}\xspace}
\newcommand{\langid}{\texttt{langid}\xspace}

\title{Language Identification with a Reciprocal Rank Classifier}

\ifcl
\author{Dominic Widdows \thanks{Email: dwiddows@liveperson.com}}
\affil{\LP Inc.} 
\author{Chris Brew \thanks{Email: cbrew@liveperson.com}}
\affil{\LP Inc.}
\runningtitle{Language Identification with a Reciprocal Rank Classifier}
\runningauthor{Widdows \& Brew}
\else
\author{Dominic Widdows\\ \texttt{dwiddows@liveperson.com} 
\and Chris Brew \\ \texttt{cbrew@liveperson.com}} 
\fi

\date{}
\begin{document}
\selectlanguage{english}

\maketitle

\begin{abstract}
\selectlanguage{english}

Language identification is a critical component of language processing pipelines
\citep{Jauhiainen2019AutomaticLI} and is not a solved problem in real-world settings.
We present a lightweight and effective language identifier that is robust to changes of 
domain and to the absence of copious training data.

The key idea for classification is that 
the reciprocal of the rank in a frequency table makes an effective additive 
feature score, hence the term Reciprocal Rank Classifier (RRC). The key finding for
language classification is that ranked lists of words and frequencies of characters form 
a sufficient and robust representation of the regularities of key languages and their orthographies.

We test this on two 22-language data sets and demonstrate zero-effort domain adaptation from
a Wikipedia training set to a Twitter test set.
When trained on Wikipedia but applied to Twitter
the macro-averaged F1-score of a conventionally trained SVM
classifier drops from 90.9\% to 77.7\%. By contrast, the
macro F1-score of RRC drops only from 93.1\% to 90.6\%. These classifiers are compared
with those from \fasttext \citep{Bojanowski2017EnrichingWV} and \langid \citep{lui-baldwin-2012-langid}. 
The RRC performs better than these established systems in most experiments, especially on short Wikipedia texts
and Twitter. An ensemble classifier that uses \fasttext when RRC abstains also performs competitively in all experiments.

The RRC classifier can be improved for particular domains and conversational situations
by adding words to the ranked lists.
Using new terms learned from such conversations,
we demonstrate a further 7.9\% increase in accuracy of sample message classification, 
and 1.7\% increase for conversation classification. Surprisingly, this made results on Twitter data
slightly worse.

The RRC classifier is available as an open source Python package.\footnote{See \url{https://github.com/LivePersonInc/lplangid}, MIT license.}
\end{abstract}

\section{Introduction}

This paper describes a new language identification system, designed for easy use and adaptation to real-world 
settings including short informal texts in conversational domains. 
The key intuition is that individual word- and character-based features are particularly robust
to domain-variation and informal use-cases. Texts are therefore classified based on the relative frequency of
their characters, and the frequency rank of each word. This makes it easy to add domain-specific 
support for new languages by taking initial text examples from a public data source such as Wikipedia, and 
adding extra terms to the word-ranking table for that language (manually or automatically) as needed.

Language identification has been studied computationally since the 1960s, leading to many 
successful methods especially based on characters, tokens, and n-grams of constituents. An early and still
canonical example of such an n-gram-based system is that of \citet{cavnar1994ngram}, which has been made
available in packages such as the implementation of \texttt{TextCat} in R \cite{feinerer2013textcat}.
A thorough analysis, comparison and history of these methods is presented in \citet{Jauhiainen2019AutomaticLI}. 

\citet{Jauhiainen2019AutomaticLI} also note that results for the same method vary considerably between datasets:
for example, \citet{lui-baldwin-2011-cross} found that the accuracy of the \texttt{TextCat} method over some test
datasets to be in the 60's, much lower than the the 99.8\% accuracy originally found by \citet{cavnar1994ngram}.
This problem is often exacerbated in commercial and customer-facing applications, where text from web-based 
and smartphone users is very unlike text from traditional training texts such as newswire and government publications
\citep{han-baldwin-2011-lexical}. As the practical and commercial uses of NLP technologies grows, this problem
has become increasingly familiar: many developers are used to trying systems whose published accuracy on curated datasets
is world-beating, but whose results on user-generated text turns out to be disappointing.

Observations like this led us to design and implement a new classifier: not because it was expected to be
better than all the others, but because we saw problems with informal conversational data and needed
a classifier that makes it easy to diagnose and fix these problems. 



\section{The Reciprocal Rank Classifier}
\label{sec:rrc}

This section explains the design and implementation of the new Reciprocal Rank Classifier. 


%
\subsection{Word Rank Scoring}

The intuition behind the new classifier is that language identification based on word-recognition is
quite robust even in informal texts. Humans who know even a little English
recognize {\it ``Tks here now''} as English because {\it here} and {\it now} are common English words,
and those who know even a little Indonesian or Malay recognize {\it ``Trmks sudah sampai''} for similar reasons.
In a sample of Wikipedia English data, the 50 most frequent words account for 37\% of the tokens, and 258 words 
is enough to cover more than half of them.

The data structure used to implement this intuition is a list of strings from each language, ranked 
by their significance. Significance can be
approximated by frequency in some training data set, 
or, if desired, manually configured according to another criterion (that this arbitrariness is a potential advantage may surprise readers in 2021, and will be discussed below).
Each text $t$ made of words $w_1 w_2 \ldots w_n$ is given a word-based score \texttt{WS} for a language $L$ as follows:
\begin{equation}
\mathtt{WS}(t, L) = \sum_{i=1}^{n} \left( P + \frac{1}{\sqrt{D + \mathtt{rank}(w_i, L)}} \right)
\label{eq:word_score}
\end{equation}
where $P$ is a term presence weight set to 0.05 (the score is incremented by this amount every time a word shows up), 
and $D$ is a damping factor set to 10 (this prevents terms from the top of the list having outsize effects).

The appearance of the rank as a term in the denominator of a scoring function gives rise to the term {\it reciprocal rank}.
The idea itself is simple and commonplace, and has been used in metrics for results ranking in information retrieval 
\cite{chapelle2009expected} and person re-identification in image processing \cite{wu2011optimizing}. While we 
have not found prior examples of using reciprocal ranking directly in scoring for a classification problem,
the idea is obvious in hindsight.

The use of reciprocal ranking of {\it words} presupposes that we know what a word is, which implies the
use of some form of tokenization. This is a well-known step in language processing, 
but as \citet{Jauhiainen2019AutomaticLI} point out:
\begin{quote}The determination of the appropriate word tokenization strategy for a given document
presupposes knowledge of the language the document is written in, which is exactly
what we assume we don’t have access to in LI.
\end{quote}
Despite this impasse, we clearly need {\it some} method for splitting texts into individual word units for the inputs
to the sum in Equation \ref{eq:word_score}. 
For this classifier implementation, the following 
tokenization rules were used:
\begin{itemize}
    \item Remove all substrings within angle brackets. (There may be examples where this is a mistake, but in our data sets,
    all such texts were found to be a source of bugs caused by treating HTML and XML tags as human language.)
    \item Replace punctuation characters with white space, except for period / full-stop characters or apostrophes
    surrounded by alphabetic characters. (For example, {\it U.S.A} $\rightarrow$ {\it U.S.A} and {\it `Rick's'} $\rightarrow$ {\it Rick's})
    \item Split the resulting string on white space.
\end{itemize}

Additionally, words containing digits (Hindu-Arabic numerals) are ignored in word-scoring, and all words are normalized to
lower-case. These decisions were made as part of error-analysis and debugging on conversational data: we do not expect these
to be universally correct policies, and have made sure they are easy to change.

To begin with, word frequencies and their corresponding ranks were collected from Wikipedia data.
Estimates of how many words people know are given by \citet{brysbaert2016many},
approximately 42K being the headline figure.
We truncated the word frequency files to use just the top 5000 words for each language, because using
more than this had no noticeable effect on performance.

\subsection{Character Frequency Scoring}
\label{sec:char_scoring}

The first version of the reciprocal rank classifier was applied to European languages using only words for scoring.
This worked badly as soon as the technique was tried for Chinese, Japanese, and Thai.
The obvious reason is that these languages do not split words using whitespace, so for example, the sentence {\it ``Chinese 
is not English''} with 3 inter-word spaces in English translates into Chinese as 
\begin{CJK*}{UTF8}{gbsn} ``中文不是英文'' \end{CJK*} 
with no spaces at all in Chinese. More quantitatively, with the whitespace tokenization above, the average `word-length'
was 8.2 characters for English, for 50.6 for Chinese, 56.9 for Japanese, and 33.0 for Thai. Of course, we should not expect 
a new text to match with these long character strings exactly.

However, each of these languages can easily be recognized by humans
because of their distinct character sets, and character-based features are widely used in language identification \citep[\S 5.2]{Jauhiainen2019AutomaticLI}. Therefore character frequencies were also gathered from Wikipedia and used
for scoring. This time, relative frequencies were used directly instead of ranks: a (normalized) frequency table is used to 
estimate the probability $P(\mathit{char}|\mathit{lang})$, this is inverted to give $P(\mathit{lang}|\mathit{char})$, and these probabilities are summed over 
the string to give an estimated $P(\mathit{lang}|\mathit{text})$. These probability scores are summed rather than multiplied to avoid 
the need for smoothing or some other way to directly model the occurrence of new characters such as emoji, and because
characters that do not `belong' to a language still occur (for example, the appearance of 
\begin{CJK*}{UTF8}{gbsn} ``中文'' \end{CJK*} just above in this document!).

Initially, character- and word- based scores were simply multiplied to give a final score for each candidate language
for a given text. Later, a character-based cutoff was introduced, to remove languages with less than $\frac{3}{4}$
of the winning character score. This was done because:
 
\begin{itemize}
    \item Wikipedias for all languages contained Roman characters, but those with a preponderance of non-Roman characters
    should not be chosen because of a few word-level matches.
    \item The use of Chinese characters in Japanese kanji led the classifier to judge that Japanese phrases with 
    Chinese characters could be Chinese, even though the appearance of Japanese-only hiragana and katakana characters
    should be enough to reject this hypothesis.
\end{itemize}

The above method has been found to work well in practice, so while 
other probabilistic and logical strategies could also be used to avoid such misclassification, we have not 
investigated these.

\subsection{Curating Words in the Ranked Lists}
\label{sec:curation}

Language detection performance varies across domains, and work on transfer learning between domains
has shown that cross-domain training is important and can give unexpected results \cite{lui-baldwin-2011-cross}.
The reciprocal rank classifier is particularly simple to configure and test,
making cross-domain behavior very easy to understand and improve, either manually or automatically.
For example, consider the English words {\it yes} and {\it thanks}. They are conversationally vital, learned early on by
anyone studying English, and each occurs in more that 1.5\% of messages in \LP conversations (sampled naively across languages).
However, they rarely occur in Wikipedia or other third-person `factual' writing: there are more than 5000 words more frequent
than {\it thanks}, and more than 14000 words more frequent than {\it yes}. So based on purely Wikipedia training data with 
a cutoff of 5000 words per language, over 3\% of English messages could be affected just because of lacking these two words!

The simplest way to add a single word for a language is to write it at the top of the language's word rank file. This was done
early on for words like {\it thanks} in English, {\it obrigado} and {\it obrigada} in Portuguese, and 
\begin{CJK*}{UTF8}{gbsn} 谢谢\end{CJK*} (Xièxiè) in Chinese, all of which mean {\it thank you}. Then to preserve these 
overrides when data is refreshed, they were checked in to a \texttt{data\_overrides} file. This was expanded to include
`computerese' stopwords --- for example, as a bug-fix, tokens starting with {\it http} are completely ignored. 

The more general and automated use of this configuration process was to add words found in \LP conversations shared by co-operating commercial customers,
essentially bootstrapping the classifier.
For example, the words {\it yes} and {\it thanks} both appear in the message ``Yes they are thanks'' (observed in practice). 
The words {\it they} and {\it are} were already known from the Wikipedia training data (rank 46 and 16 respectively),
so the classifier easily classifies ``Yes they are thanks'' as English, which is a (weak) indication that {\it yes} and {\it thanks}
are English words as well. Over a large collection, such statistics can be collected, giving a list of which words most commonly
appear in English messages in spite of being unknown to the English classifier. Table \ref{table:english_words} shows
the most frequent words found in this way, and in particular, the last column shows the most frequent words in \LP English 
conversations that were {\it not} frequent enough in Wikipedia to occur in the list of top 5000 words. Some are clearly domain-specific and 
related to customer service and contact ({\it email, payment, chat, wait, customer}), whereas some are typical of first- and second-person
conversational utterances that are poorly represented in Wikipedia ({\it please, thank, hi, yes}). The word {\it no} has been included in this
table as a contrast to {\it yes} --- the word {\it no} is still common in Wikipedia, because it appears in more settings 
than {\it yes} (contrast ``This is no problem'' with the ungrammatical ``This is yes problem'').

Classification results were good when using the most frequent 100 words from the conversational data for each language, provided 
they did not already occur in the ranking table. The lists were merged by making $k^{\mathrm{th}}$ term in the conversation
data the $k^{\mathrm{th}}$ term in the merged ranking table, moving the subsequent Wikipedia terms one place further down in the list.
New conversational terms that failed the character cutoff test mentioned in Section \ref{sec:char_scoring} were rejected.
The overall point is that \rrc makes deficiencies in the training material both discoverable and actionable.

\begin{table}
\begin{center}
\begin{small}

\begin{tabular}{ccc}

\begin{tabular}{|c|}
    \hline
    \makecell{Top 10 \\ Wikipedia} \\ 
    \hline
of  \\
and \\
in  \\
to  \\
is  \\
the  \\
as   \\
was \\
that \\
for  \\
\hline
\end{tabular}

&

\begin{tabular}{|c|}
    \hline
    \makecell{Top 10 \\ \LP} \\ 
    \hline
you \\
to  \\
the \\
I \\
your  \\
a  \\
for  \\
and \\
what \\
is  \\
\hline
\end{tabular}

&

\begin{tabular}{|c|c|c|}
    \hline
\makecell{High in \LP \\ Low in Wikipedia} & \makecell{\LP \\ Rank} & \makecell{Wikipedia \\ Rank} \\ 
    \hline
please & 11 & 9969 \\
thank & 37 & 16565 \\
hi & 52 & 9688 \\
yes & 53 & 8278 \\
\textit{no} & \textit{41} & \textit{70} \\
email & 81 & 7029 \\
payment & 97 & 5312 \\
chat & 104 & 12086 \\
wait & 140 & 7018 \\
customer & 116 & 6172 \\
\hline
\end{tabular}

\end{tabular}
\end{small}
\caption{Comparing the most frequent words in English Wikipedia and \LP Conversations}
\label{table:english_words}
\end{center}
\end{table}

The explicit and editable format of these files also enables easy additions and deletions in response
to error reports and direct customer requests. 
The ability to support such operations quickly and easily, with no costly retraining, contributes to better user experiences 
(particularly in removing unanticipated bugs), immediate verification of fixes, and more cordial
collaboration between science and product teams in a commercial organization. This a deliberate design choice, and 
is part of the general issue of model interpretability in machine learning: the reciprocal rank model is
additive (as described by \citet{rudin2019stop}) and this simplicity contributes to behavior that is
easy to explain and improve.

\section{Evaluation with Public Data}
\label{sec:evaluation}

This section presents quantitative evaluation results on classifying different language samples drawn
from Wikipedia and Twitter datasets, focusing on 22 languages. The \rrc is evaluated and compared with 
publicly available classifiers (\fasttext and langid), and another machine learning classifier (SVC) built
for comparison using the same Wikipedia training data as the RRC.

We show that the RRC and SVC classifiers perform well and quite similarly when trained and evaluated on Wikipedia.
RRC also does better than the publicly available classifiers on several of the tasks, especially  
when restricted to 16 characters. RRC achieves the best results with most languages on the 
Twitter dataset, and is considerably better than SVC here, even though both were trained on the same Wikipedia data.
These results demonstrate that the RRC provides comparable results to state-of-the-art classifiers on easy tasks,
and improves on the state-of-the-art for some more difficult tasks.

\subsection{Materials Used in Experiments}

\subsubsection{Datasets and Languages}
\label{sec:datasets}

We use two sources of generally available multilingual data.

\paragraph{Wikipedia}\citep{wiki:dumps} 

Wikipedia was used for training and within-dataset evaluation. 
    The are 56,175 files in the Wikipedia dumps for our target languages. English is the largest, with 15,002 files, and Tagalog the smallest, with just 62. We split the files 80\%:20\% train:test.
    We randomly selected up to 20 files from each language for use as training material, and another 20 (if available) for testing. 
    This yields a training corpus with 1,201,891 total lines. 
    The main results reported below use a stratified sample of 10k Wikipedia samples with a target sample size of 256 (see Section~\ref{sample:size} for the details of the sampling process).


All Wikipedias were found to include substantial numbers of English-language names and phrases, as 
well as English language material associated with bibliography. So when we see an `error' that involves English, we can be 
appropriately skeptical. It could really be a sample that really is predominantly or entirely in English, even though it appears in the
Wikipedia for another language. 

\paragraph{Twituser} \citep{lui-baldwin-2014-accurate}. 

The \texttt{twituser} dataset was used only for evaluation. For the classifiers trained only on Wikipedia, these results also measure the effectiveness of zero-shot transfer from the Wikipedia domain.
\texttt{Twituser} contains 14,178 tweets, of which 8910 are assigned to one of our 22 languages. As was mentioned for Wikipedia, `assigned' is appropriately careful: not all the assignments are correct, and in many cases it
is impossible to come to a definitive judgment about which language the tweet is `really` in.

For example:
\begin{itemize}
    \item \texttt{dreaming} is assigned to Chinese.
    \item  \texttt{Eeyup!} is assigned to German\footnote{It is dialect English, probably used mostly for comedic effect.}.
\end{itemize}
We clean the tweets using the method provided by \citep{lui-baldwin-2014-accurate}, which is in turn inherited from \texttt{twokenize}\footnote{\url{https://github.com/leondz/twokenize}}.

\paragraph{Languages}

The languages we tested in these experiments are listed in Table~\ref{target:languages}. 
These languages were particularly pertinent to our development work:
the github package already includes a few more and a \texttt{bash} script for
automatically adding new languages from recent Wikipedia data.

\begin{table*}
\caption{Target languages}
    \label{target:languages}
\center
\small
\begin{tabular}{ll}
\toprule
ISO 639-1 &   English name           \\
\midrule
ar        &       Arabic \\
de        &       German \\
el        &        Greek \\
en        &      English \\
es        &      Spanish \\
fr        &       French \\
he        &       Hebrew \\
hi        &        Hindi \\
id        &   Indonesian \\
it        &      Italian \\
ja        &     Japanese \\
\bottomrule
\end{tabular}
\quad
\begin{tabular}{ll}
\toprule
ISO 639-1 &    English name         \\
\midrule
ko        &       Korean \\
mk        &   Macedonian \\
nl        &        Dutch \\
pt        &   Portuguese \\
ru        &      Russian \\
sl        &    Slovenian \\
sq        &     Albanian \\
th        &         Thai \\
tl        &      Tagalog \\
vi        &   Vietnamese \\
zh        &      Chinese \\
\bottomrule
\end{tabular}

\end{table*}

\subsubsection{Classifiers}
\label{sec:classifiers}

We measure performance for the following language classifiers:

\begin{itemize}
    \item \texttt{RRC} The Reciprocal Rank Classifier. This is ``trained'' by counting words and characters in Wikipedia 
    dump files for the relevant languages, and adding words for conversational data described in Section~\ref{sec:curation}.
    A total of 308 words across 11 languages were added in this way.

    \item \texttt{SVC} A linear support vector classifier implemented in \texttt{scikit-learn} \citep{scikit-learn}. This uses the same features as the \rrc (case-lowered word unigrams and case-preserved character unigrams), and relies on the \texttt{HashingVectorizer} to achieve efficient, stateless vectorization. 
    The regularization parameter, \texttt{C} was set to $2.3$  using \texttt{scikit-learn}'s standard \texttt{GridSearchCV} method. Otherwise, the default parameters were used.
    
    
    \item \texttt{langid.py} \citep{lui-baldwin-2012-langid} using the publicly available model, but set to consider only our 22 target languages.
    \item \texttt{fasttext}'s out-of-the-box 176-language identifier \citep{Bojanowski2017EnrichingWV}, with a wrapper that restricts to our 22 languages.
\end{itemize}

The addition of conversational words to the RRC is a curated improvement, and whether this is a fair advantage over the SVC 
classifier depends on whether it is regarded as a difference in design or training data. We would have added conversational words to the SVC
if its model supported this, and the \fasttext and \langid classifiers were used with no restrictions on how they were trained, so
we reckoned that the comparison was fair. In the event, this debate turned out to be splitting hairs: the RRC got the best individual results
on the more conversational Twitter dataset with and without the help of the \LP conversational words (see Section \ref{sec:twitter_results}).

\subsection{Wikipedia Results for New Classifiers}

This section summarizes classification results on the Wikipedia dataset described in Section~\ref{sec:datasets}. 
The task is as follows:
given a selection of text drawn from a Wikipedia file from a language-specific wikidump,
say which language's wikidump it was drawn from.

In order to simulate realistic task conditions, we tested samples of various target sizes, extracted 
from the Wikipedia test corpus.
We do not want to make samples where a word is artificially broken in two, so
the samples for a target size of $k$ are defined to be the non-overlapping chunks obtained by greedily going through the text from start to finish such that each chunk:
\begin{itemize}
    \item is white-space-delimited.
    \item has a total length $>= k$ unless it is the final chunk.
    \item does not contain any extra words after the target has been reached.
\end{itemize}
The final chunk of each file may be shorter than the target size. All others meet or exceed the target size.

The first experiment we report was performed on texts of target size 256. The experiment
compares the results of the SVC and RRC classifiers that were built only using the training
part of this Wikipedia dataset. We use balanced F1 score as the main metric of performance.
Results are given in Table \ref{tab:wikipedia_sgd_rrc_report}.

\begin{table}
\caption{Classifier F1 Scores on Wikipedia dataset for SVC and RRC. Sample length 256.}
\label{tab:wikipedia_sgd_rrc_report}
\renewcommand{\arraystretch}{1.15}
\begin{small}
\begin{center}
\begin{tabular}{|lr|rrr|rrr|}
\hline
Language     & Support  & \multicolumn{3}{c|}{SVC} & \multicolumn{3}{c|}{RRC} \\
{} & {} & Precision & Recall & F1-score &  Precision & Recall & F1-score \\

\hline

ar           &      351 &  98.86  &  98.58  &  98.72  &  99.43  &  98.86  &  \bf{99.14}  \\
de           &      447 &  96.97  &  93.06  &  \bf{94.98}  &  94.82  &  94.18  &  94.50  \\
el           &      290 &  99.29  &  96.21  &  97.72  &  99.64  &  96.55  &  \bf{98.07}  \\
en           &      369 &  90.98  &  92.95  &  \bf{91.96}  &  79.91  &  97.02  &  87.64  \\
es           &      401 &  93.48  &  93.02  &  93.25  &  93.38  &  95.01  &  \bf{94.19}  \\
fr           &      525 &  96.47  &  93.71  &  \bf{95.07}  &  88.03  &  95.24  &  91.49  \\
he           &      376 &  100.00  &  99.47  &  99.73  &  100.00  &  99.73  &  \bf{99.87}  \\
hi           &      221 &  99.54  &  98.64  &  99.09  &  99.55  &  99.55  &  \bf{99.55}  \\
id           &      656 &  78.62  &  81.86  &  80.21  &  96.90  &  80.95  &  \bf{88.21}  \\
it           &      652 &  96.27  &  83.13  &  \bf{89.22}  &  93.24  &  84.66  &  88.75  \\
ja           &      474 &  99.54  &  90.51  &  94.81  &  99.54  &  91.56  &  \bf{95.38}  \\
ko           &      652 &  98.49  &  90.18  &  94.16  &  99.84  &  97.09  &  \bf{98.44}  \\
mk           &      366 &  98.51  &  90.16  &  \bf{94.15}  &  99.69  &  88.80  &  93.93  \\
nl           &      585 &  44.09  &  84.10  &  57.85  &  95.88  &  75.56  &  \bf{84.51}  \\
pt           &      490 &  96.10  &  90.61  &  93.28  &  97.42  &  92.45  &  \bf{94.87}  \\
ru           &      278 &  95.37  &  96.40  &  95.89  &  99.63  &  95.68  &  \bf{97.61}  \\
sl           &      623 &  86.99  &  85.87  &  86.43  &  97.23  &  84.59  &  \bf{90.47}  \\
sq           &      549 &  99.31  &  78.51  &  87.69  &  98.90  &  81.60  &  \bf{89.42}  \\
th           &      267 &  100.00  &  96.63  &  98.29  &  100.00  &  97.38  &  \bf{98.67}  \\
tl           &      365 &  87.88  &  71.51  &  78.85  &  91.07  &  72.60  &  \bf{80.79}  \\
vi           &      535 &  87.98  &  80.75  &  84.21  &  96.83  &  80.00  &  \bf{87.62}  \\
zh           &      528 &  97.79  &  92.23  &  94.93  &  98.65  &  96.97  &  \bf{97.80}  \\
\hline
accuracy     &        0 &  88.75  &  88.75  &  88.75  &  89.58  &  89.58  &  \bf{89.58}  \\
macro avg    &    10000 &  92.84  &  89.91  &  90.93  &  96.17  &  90.55  &  \bf{93.05}  \\
weighted avg &    10000 &  91.55  &  88.75  &  89.60  &  96.23  &  89.58  &  \bf{92.55}  \\

\hline

\end{tabular}
\end{center}
\end{small}
\end{table}

The RRC classifier does better on 17 of the 22 languages, and achieves a better accuracy
(total proportion of correct answers), macro average F1 score (unweighted by class size), 
and weighted average F1 score (weighted by class size). 
The RRC results are particularly strong for the languages with distinct character sets: the results for 
Arabic, Greek, Hebrew, Hindi, Korean, and Thai are all very high. 
Neither classifier does particularly well for English, and the precision for the RRC
at 79.9\% is particularly weak. This is partly expected due to the appearance of English words in other Wikipedias noted above.
In spite of the disproportionately strong support for English in NLP tools and resources, 
for the language classification task, the prevalence of English makes it {\it harder}
to classify correctly.

\begin{table}[h]
\begin{center}
\caption{Summary results for Wikipedia task for four classifiers. Sample length 256.}
\label{tab:wikipedia_summary_256}
\begin{tabular}{lrrrr}
\toprule
{} & fasttext & langid & SVC & RRC \\
\midrule
accuracy        &  \bf{90.37}  &  88.99  &  88.75  &  89.58  \\
macro avg precision    &  93.00  &  92.02  &  90.93  &  \bf{96.17}  \\
weighted avg precision &  93.29  &  92.30  &  92.84  &  \bf{96.23}  \\
macro avg recall    &  \bf{91.53}  &  90.26  &  89.91  &  90.55  \\
weighted avg recall &  \bf{90.37}  &  88.99  &  88.75  &  89.58  \\
macro avg F1    &  91.36  &  90.14  &  90.93  &  \bf{93.05}  \\
weighted avg F1 &  90.98  &  89.71  &  89.60  &  \bf{92.55}  \\
\bottomrule
\end{tabular}
\end{center}
\end{table}

Table~\ref{tab:wikipedia_summary_256} compares these results with those of the other two established classifiers. 
The differences are modest:
the performance of the SVC and RRC classifiers trained here is comparable 
with the state-of-the-art on this task, \fasttext gets the highest accuracy, 
and the RRC's macro and weighted averages are the best. However, the new classifiers were trained and evaluated on Wikipedia
data, so they might be overfitted to this task, a hypothesis explored below in Section \ref{sec:twitter_results}.

\subsection{Wikipedia Results for Short Inputs}
\label{sample:size}

Conversational applications include many short messages for which a unique language classification is impossible in isolation.
An obvious example is that the single word message ``No'' occurs relatively often, which could be a negative response
to a question in at least English, Spanish, and Italian. However much training data we amassed for any one of these languages,
any statistical claim that ``No'' is most likely to be (say) Spanish would reflect only the bias of the training data,
rather than the language used in any particular conversation. To know what language is being used in a particular
question / answer pair, we need to know what question was asked. 
For example, ``¿Quieres algo más? No.'' is Spanish and ``Do you want anything else? No.'' is English. 

Systems that perform well on long curated texts such as news articles or parliamentary records
sometimes give poor outcomes with informal text messages.
This vulnerability is noted in previous work:

\begin{quote}
The size of the input text is known to play a significant role in the accuracy of automatic language
identification, with accuracy decreasing on shorter
input documents. \linebreak 
\vspace{-0.2in}
\flushright{\citep{lui-baldwin-2012-langid}}
\end{quote}

Thus we expected classifier results to be worse with shorter messages. 
We evaluated the two new classifiers (SVC and RRC) and the \fasttext and \langid classifiers in this way. 
Results of this experiment for texts of lengths 16 and 64 are shown in Table~\ref{tab:length_results}. 
As expected, each classifier loses performance compared with the results on longer texts in Table~\ref{tab:wikipedia_summary_256}.
However, they are not affected equally: the performance of \fasttext and RRC remains relatively strong compared with
langid and SVC, which drop more steeply. \fasttext and \langid also become particularly weak at classifying English.
Given that short English messages are quite prevalent in many applications, the relative weakness of English classification
of short texts is something developers should watch out for, and for which the \rrc is a comparatively good choice.

\begin{table}[t]
\caption{Classifier F1 scores for samples of length 16 (left) and 64 (right). Performance degrades with the shorter test samples. 
\fasttext and RRC degrade the least.}
\label{tab:length_results}
\begin{adjustbox}{width=\linewidth}
\begin{tabular}{c}

\footnotesize
\begin{tabular}{lrrrrrrr}
\toprule\multicolumn{2}{r}{fasttext} &  langid  &  SVC  &  RRC\\
\midrule
ar           &  95.40  &  98.14  &  98.13  &  \bf{98.71}  \\
de           &  82.60  &  73.78  &  75.92  &  \bf{85.65}  \\
el           &  96.81  &  \bf{96.84}  &  96.65  &  \bf{96.84}  \\
en           &  47.49  &  40.85  &  64.59  &  \bf{77.50}  \\
es           &  76.66  &  62.38  &  67.60  &  \bf{79.39}  \\
fr           &  78.13  &  73.22  &  75.72  &  \bf{83.26}  \\
he           &  98.93  &  98.93  &  98.52  &  \bf{99.20}  \\
hi           &  98.17  &  \bf{99.55}  &  97.93  &  99.32  \\
id           &  77.14  &  66.60  &  65.59  &  \bf{82.26}  \\
it           &  79.11  &  70.19  &  70.80  &  \bf{81.13}  \\
ja           &  \bf{95.42}  &  91.95  &  92.62  &  91.70  \\
ko           &  97.58  &  \bf{98.37}  &  92.52  &  97.65  \\
mk           &  \bf{92.24}  &  84.78  &  84.57  &  89.25  \\
nl           &  \bf{77.03}  &  71.91  &  38.97  &  75.86  \\
pt           &  81.17  &  73.65  &  69.90  &  \bf{83.31}  \\
ru           &  \bf{89.33}  &  83.53  &  84.72  &  81.86  \\
sl           &  77.74  &  74.74  &  69.37  &  \bf{81.51}  \\
sq           &  78.64  &  76.65  &  73.92  &  \bf{83.84}  \\
th           &  \bf{96.51}  &  \bf{96.51}  &  95.29  &  \bf{96.51}  \\
tl           &  65.21  &  47.77  &  69.42  &  \bf{75.75}  \\
vi           &  \bf{84.74}  &  83.93  &  72.04  &  84.53  \\
zh           &  95.00  &  91.32  &  92.47  &  \bf{95.22}  \\
\midrule
accuracy     &  \bf{82.64}  &  77.43  &  74.97  &  81.82  \\
macro avg    &  84.59  &  79.80  &  79.42  &  \bf{87.12}  \\
weighted avg &  83.78  &  78.78  &  77.27  &  \bf{86.47}  \\
\bottomrule
\end{tabular}
\quad
\footnotesize
\begin{tabular}{lrrrrrrr}
\toprule\multicolumn{2}{r}{fasttext} &  langid  &  SVC  &  RRC\\
\midrule
ar           &  96.79  &  98.72  &  98.86  &  \bf{99.14}  \\
de           &  88.43  &  83.88  &  92.72  &  \bf{93.57}  \\
el           &  \bf{97.54}  &  97.20  &  97.02  &  97.20  \\
en           &  57.34  &  53.11  &  \bf{86.80}  &  85.78  \\
es           &  87.29  &  81.06  &  89.23  &  \bf{91.39}  \\
fr           &  86.94  &  83.62  &  \bf{90.22}  &  89.77  \\
he           &  99.73  &  99.47  &  99.47  &  \bf{99.87}  \\
hi           &  98.41  &  \bf{99.55}  &  98.63  &  \bf{99.55}  \\
id           &  85.45  &  80.21  &  77.28  &  \bf{86.58}  \\
it           &  85.96  &  82.16  &  86.03  &  \bf{86.68}  \\
ja           &  \bf{96.87}  &  94.83  &  94.22  &  94.22  \\
ko           &  98.13  &  \bf{98.84}  &  94.08  &  98.44  \\
mk           &  \bf{94.25}  &  90.54  &  90.75  &  90.94  \\
nl           &  \bf{84.71}  &  81.91  &  55.59  &  81.75  \\
pt           &  91.02  &  89.87  &  88.58  &  \bf{93.78}  \\
ru           &  \bf{92.70}  &  89.80  &  91.65  &  92.66  \\
sl           &  88.83  &  86.37  &  83.27  &  \bf{88.91}  \\
sq           &  87.60  &  85.92  &  86.01  &  \bf{88.08}  \\
th           &  \bf{98.67}  &  \bf{98.67}  &  97.50  &  \bf{98.67}  \\
tl           &  77.46  &  67.59  &  78.02  &  \bf{80.00}  \\
vi           &  \bf{88.03}  &  87.68  &  82.41  &  87.18  \\
zh           &  96.41  &  94.35  &  94.26  &  \bf{97.02}  \\
\midrule
accuracy     &  \bf{88.80}  &  86.17  &  86.32  &  87.75  \\
macro avg    &  89.94  &  87.52  &  88.75  &  \bf{91.70}  \\
weighted avg &  89.53  &  87.01  &  87.35  &  \bf{91.19}  \\
\bottomrule
\end{tabular}
\end{tabular}
\end{adjustbox}
\end{table}

\subsection{Abstentions and Ensemble Classification}
\label{sec:abstentions_ensemble}

The \rrc abstains when it is unsure, whereas the other classifiers make a forced choice.
Abstention happens if a text is all digits, or has recognized characters for several languages but no recognized words.
A few examples from short messages (from the 16-character Wikipedia test sample) are shown in Table~\ref{tab:abstentions}.
For some of these texts, assigning them to a language is clearly a mistake (unless that language is HTML!). Some are clearly
`human to human messages', but international (these include digits and scientific names drawn from Latin). Some are classifier mistakes:
for example, {\it Wissenschaftseinrichtungen} is the German for `scientific institutions', and the failure of the RRC to
recognize demonstrates a natural weakness of using whitespace-delimited `words' as features with languages that use
a lot of compound words.

\begin{table}
\begin{center}
\renewcommand\arraystretch{1.5}
\caption{Short messages where the \rrc abstains}
\label{tab:abstentions}
\begin{small}
\begin{tabular}{lclcl}
EON www.eon.tv & & Wissenschaftseinrichtungen & &</br> \\
Dannebrog & & \begin{CJK*}{UTF8}{gbsn} 目次へ移動 \end{CJK*} & & 1906 \\
1962 & & sistim Gereja – Gereja & & 5 \\
Eimeria augusta & & Navarone Foor & & analgesik inhalasi\\
2 13 0.01101 2 5 & & <section begin=Pekan 29 /> & & rVSV SUDV  \\
\end{tabular}
\end{small}
\end{center}
\end{table}

Abstention allows a simple form of \textbf{ensemble classification}, in which another classifier (in our case out-of-the-box \fasttext) is used to provide results for the samples on which the \rrc abstains. Results for this classifier on the Wikipedia 16 and 256 character samples are given
in Table~\ref{tab:ensemble_results}. For longer texts, comparatively little changes, but for the short texts, the ensemble classifier 
does the best for most languages and builds a 3\% lead in overall accuracy over either of the ingredients. 

\begin{table}[t]
\caption{Ensemble classification compared with individual classifier F1 scores for samples of length 16 (left) and 256 (right)}
\label{tab:ensemble_results}
\begin{adjustbox}{width=\linewidth}
\begin{tabular}{c}
\footnotesize
\begin{tabular}{lrrr}
\toprule\multicolumn{2}{r}{fasttext} &  RRC  &  Ensemble\\
\midrule
ar           &  95.40  &  \bf{98.71}  &  96.90  \\
de           &  82.60  &  \bf{85.65}  &  84.47  \\
el           &  96.81  &  \bf{96.84}  &  \bf{96.84}  \\
en           &  47.49  &  \bf{77.50}  &  57.27  \\
es           &  76.66  &  \bf{79.39}  &  76.92  \\
fr           &  78.13  &  \bf{83.26}  &  78.90  \\
he           &  98.93  &  \bf{99.20}  &  98.93  \\
hi           &  98.17  &  \bf{99.32}  &  \bf{99.32}  \\
id           &  77.14  &  82.26  &  \bf{82.79}  \\
it           &  79.11  &  \bf{81.13}  &  80.56  \\
ja           &  95.42  &  91.70  &  \bf{96.27}  \\
ko           &  97.58  &  97.65  &  \bf{98.13}  \\
mk           &  92.24  &  89.25  &  \bf{94.38}  \\
nl           &  77.03  &  75.86  &  \bf{78.97}  \\
pt           &  81.17  &  \bf{83.31}  &  82.71  \\
ru           &  89.33  &  81.86  &  \bf{91.55}  \\
sl           &  77.74  &  81.51  &  \bf{84.14}  \\
sq           &  78.64  &  83.84  &  \bf{84.62}  \\
th           &  \bf{96.51}  &  \bf{96.51}  &  \bf{96.51}  \\
tl           &  65.21  &  75.75  &  \bf{75.82}  \\
vi           &  84.74  &  \bf{84.53}  &  84.47  \\
zh           &  95.00  &  95.22  &  \bf{96.23}  \\
\midrule
accuracy     &  82.64  &  81.82  &  \bf{85.77}  \\
macro avg    &  84.59  &  \bf{87.12}  &  \bf{87.12}  \\
weighted avg &  83.78  &  \bf{86.47}  &  86.39  \\
\bottomrule
\end{tabular}
\quad
\footnotesize
\begin{tabular}{lrrr}
\toprule\multicolumn{2}{r}{fasttext} &  RRC  &  Ensemble\\
\midrule
ar           &  97.06  &  \bf{99.14}  &  97.88  \\
de           &  90.07  &  \bf{94.50}  &  90.91  \\
el           &  \bf{98.07}  &  \bf{98.07}  &  97.90  \\
en           &  60.26  &  \bf{87.64}  &  68.76  \\
es           &  91.12  &  \bf{94.19}  &  90.80  \\
fr           &  88.25  &  \bf{91.49}  &  86.82  \\
he           &  99.60  &  \bf{99.87}  &  99.60  \\
hi           &  99.09  &  \bf{99.55}  &  \bf{99.55}  \\
id           &  87.04  &  88.21  &  \bf{88.29}  \\
it           &  87.70  &  \bf{88.75}  &  86.40  \\
ja           &  97.10  &  95.38  &  \bf{97.58}  \\
ko           &  98.13  &  98.44  &  \bf{98.60}  \\
mk           &  96.03  &  93.93  &  \bf{97.05}  \\
nl           &  87.13  &  84.51  &  \bf{86.79}  \\
pt           &  93.44  &  \bf{94.87}  &  93.67  \\
ru           &  94.83  &  \bf{97.61}  &  95.80  \\
sl           &  91.15  &  90.47  &  \bf{90.82}  \\
sq           &  88.73  &  89.42  &  \bf{89.57}  \\
th           &  98.48  &  \bf{98.67}  &  \bf{98.67}  \\
tl           &  80.39  &  \bf{80.79}  &  \bf{80.79}  \\
vi           &  88.57  &  87.62  &  \bf{87.64}  \\
zh           &  97.61  &  97.80  &  \bf{98.01}  \\
\midrule
accuracy     &  90.37  &  89.58  &  \bf{91.11}  \\
macro avg    &  91.36  &  \bf{93.05}  &  91.91  \\
weighted avg &  90.98  &  \bf{92.55}  &  91.40  \\
\bottomrule
\end{tabular}
\end{tabular}
\end{adjustbox}
\end{table}

Abstention can lead to differences in evaluation results based on how tasks are set up.\footnote{On a 
simple programmatic level, the \texttt{sklearn} function that makes classification reports breaks with any
\texttt{None} values, so even with such a standard library, the caller has to decide how to handle this.}
For example, with the Wikipedia tasks, abstentions are always marked incorrect, because even if a 
non-language text (such as one of the examples in Table~\ref{tab:abstentions}) is drawn from the 
Wikipedia for language X, X is treated as the correct label for the
text, though it could have come from several other Wikipedias. This incentivizes classifiers to guess (claiming false 
confidence) rather than to abstain (which is clearly more correct in some cases).
For macro averages, the denominator is just the number of classes, 
so if `abstention' is considered as a class in its own right, the denominator for the macro average for the RRC is 23 rather than
22, and the RRC macro average scores would be correspondingly smaller.
(The accuracy and weighted average scores are unchanged by this.) 
The RRC could be changed to give a best guess (e.g., based on characters alone), but we have avoided doing this 
because it would be an optimization for particular experimental conditions rather than a desirable feature in general.

More generally, allowing abstention is a kind of precision / recall tradeoff \cite[Ch 3]{geron2019hands}. A classifier 
that decides to `pass' when unsure is able to maintain higher precision by avoiding borderline decisions,
but if there is a correct answer for this case, the classifier is guaranteed to lose some recall.
As shown in Table~\ref{tab:wikipedia_summary_256}, the RRC does have particularly high precision, loses
recall compared to the \fasttext classifier, and according to the F1 score, makes the best combination.

\subsection{Evaluation with Twitter Data}
\label{sec:twitter_results}

The last experiment in this section compares results on the {\it Twituser} dataset. 
The new SVC and RRC classifiers had only been trained so far on Wikipedia, and
Tweets are expected to be less curated, and somewhat more like conversational text.
We compared the four classifiers from Section \ref{sec:classifiers}, 
using two versions of the RRC, `RRC Full' which includes the conversational
terms described in Section \ref{sec:curation} and `RRC Wiki' which uses only the 
term ranks from Wikipedia frequency count.
The ensemble of the
RRC Full and \fasttext classifiers introduced in Section \ref{sec:abstentions_ensemble} was also tested.
The results are in Table~\ref{tab:ensemble_results}.

The biggest casualty is the SVC classifier, whose average performance drops from the high 80's to 
the high 70's. This classifier is apparently not robust to the change of domain. The \langid classifier comes
into its own, achieving top performance in 4 languages, while \fasttext
gets top performance in 5 languages.
The ensemble is still statistically strong, with the best overall accuracy, but
doing the best in only 4 languages, Chinese, Thai, and Hindi (for which 
all classifiers score perfectly, due to the character set and the limited set of languages used in these experiments).
Some version of RRC gets top performance in 14 languages, with particularly strong performance on English and languages with distinct character sets.
The RRC classifiers 
get the highest accuracy of any of the individual classifiers, and the best macro and weighted average of all.
Surprisingly, the `RRC Wiki' term ranks {\it without} the addition of conversational words performs the best.  
As a newcomer with training data from Wikipedia and no previous exposure to Twitter data,
the \rrc holds up very well in this experiment, demonstrating the zero-effort domain adaptation 
referred to at the outset.

\begin{table}[t]
\caption{F1-scores on the {\it Twituser} dataset for all classifiers}
\label{tab:ensemble_results}
\begin{center}
\begin{tabular}{lrrrrrrr}
\toprule\multicolumn{2}{r}{support} & fasttext &  langid  &  SVC  &  \makecell{RRC\\Full}  &  \makecell{RRC\\ Wiki}  &  Ensemble\\
\midrule
ar           &      497 &  95.61  &  98.47  &  98.07  &  \bf{98.57}  &  98.47  &  98.07  \\
de           &      489 &  89.28  &  90.66  &  85.59  &  \bf{93.70}  &  92.89  &  92.05  \\
el           &      493 &  97.84  &  98.98  &  98.15  &  \bf{99.08}  &  98.77  &  98.88  \\
en           &      495 &  66.23  &  71.26  &  50.95  &  85.97  &  \bf{87.63}  &  80.39  \\
es           &      489 &  89.83  &  84.21  &  73.24  &  88.69  &  \bf{90.08}  &  89.44  \\
fr           &      492 &  88.57  &  88.47  &  81.30  &  92.79  &  \bf{93.04}  &  91.81  \\
he           &      496 &  99.49  &  98.79  &  99.29  &  \bf{100.00}  &  99.80  &  \bf{100.00}  \\
hi           &       30 &  \bf{100.00}  &  \bf{100.00}  &  \bf{100.00}  &  \bf{100.00}  &  \bf{100.00}  &  \bf{100.00}  \\
id           &      486 &  78.22  &  78.89  &  70.76  &  84.91  &  \bf{86.22}  &  85.99  \\
it           &      485 &  88.96  &  86.30  &  82.93  &  \bf{91.65}  &  90.95  &  91.45  \\
ja           &      497 &  93.76  &  97.31  &  94.03  &  98.78  &  \bf{98.89}  &  98.20  \\
ko           &      496 &  90.87  &  98.08  &  78.51  &  \bf{99.60}  &  98.78  &  99.30  \\
mk           &       52 &  \bf{90.72}  &  84.68  &  77.19  &  83.93  &  84.21  &  84.96  \\
nl           &      484 &  \bf{89.94}  &  84.04  &  56.35  &  85.46  &  89.18  &  88.31  \\
pt           &      490 &  \bf{88.37}  &  84.79  &  69.52  &  85.46  &  86.60  &  87.30  \\
ru           &      486 &  98.17  &  \bf{98.34}  &  96.52  &  92.72  &  93.54  &  98.02  \\
sl           &       58 &  \bf{85.47}  &  77.17  &  22.13  &  62.86  &  65.14  &  61.45  \\
sq           &       90 &  77.71  &  \bf{81.71}  &  48.33  &  74.63  &  71.68  &  73.89  \\
th           &      498 &  98.58  &  \bf{99.09}  &  97.64  &  \bf{99.09}  &  98.99  &  98.99  \\
tl           &      340 &  72.83  &  68.14  &  64.40  &  \bf{86.23}  &  85.88  &  85.59  \\
vi           &      478 &  93.36  &  94.26  &  74.62  &  94.23  &  93.03  &  \bf{94.36}  \\
zh           &      489 &  92.81  &  96.48  &  89.43  &  97.70  &  97.05  &  \bf{98.05}  \\
\midrule
accuracy     &        0 &  89.47  &  89.90  &  79.58  &  90.55  &  90.89  &  \bf{92.65}  \\
macro avg    &     8910 &  89.39  &  89.10  &  77.68  &  90.55  &  \bf{90.95}  &  90.75  \\
weighted avg &     8910 &  89.78  &  90.04  &  80.85  &  92.77  &  \bf{93.04}  &  92.84  \\
\bottomrule
\end{tabular}
\end{center}
\end{table}

\subsection{Conclusions from Experiments}

Taken as a whole, the experiments in this section support the following claims:

\begin{itemize}
    \item The \rrc performs well compared with state-of-the-art classifiers on all tasks, and sometimes clearly better.
    \item The RRC supports zero-effort domain transfer from from Wikipedia to Twitter. 
    The addition of conversational terms to the RRC did not improve this domain transfer.
    \item The character-based component of RRC successfully handles languages with distinctive character sets.
    \item The RRC is comparatively robust for handling short fragments of text, though results degrade for all classifiers.
    \item Strong performance is achieved by an ensemble between the RRC and \texttt{fasttext}.
\end{itemize}

The \rrc is a lightweight, robust, adaptable classifier, and its use can be recommended in many practical situations.

\section{Observations with Conversational Data}

Robust and easy adaptation to the conversational domain was a key motivation throughout this work.
This section focuses on examples
of how the design of the classifier 
affects customer service conversations.

\subsection{Messages vs Conversations}
\label{sec:messages_conversations}

Section \ref{sample:size} demonstrated an expected drop in classification accuracy for short messages.
This is a problem we had encountered in practice, and 
the effect can be quantified very simply and realistically in conversational data: conversations are lists of individual messages, 
and the messages do not all arrive at once. So how often does the language detector's prediction change {\it during} a conversation?
On samples of 50K conversations from different regions and date ranges, we found that for between 12\% and 14\% of messages, the
RRC classifier gave a different language for the individual message than for the conversation as a whole. This range was slightly
smaller for the first message in the conversation: for between 11\% and 13\%, the first message was classified differently from
the conversation as a whole.

Thus, even for a classifier reporting good results in evaluation, it was relatively easy to pick out
a large proportion of apparent errors (14\%). Some of these were subsequently fixed using the curation process described in 
Section~\ref{sec:curation}, for example, boosting the rank of the word ``hi'' in English.
However, many of these errors reflect properties of the domain (customer service live chat) rather than properties
of the classifier, or any particular language. This result exemplifies a core recommendation from this work: if an application really needs to improve the accuracy of 
its language classification step, finding a way to get longer or more informative inputs is more important than improving any classifier model.

\subsection{Words Added from \LP Conversations}
\label{sec:lp_words_added}

The relative rarity in Wikipedia of standard words like {\it Yes} and {\it Thanks} and the method for addressing this
was outlined in Section ~\ref{sec:curation}. This section summarizes the results of this process in practice.

Conversations from cooperating partners were sampled over a period of two days during October 2020.
Each conversation over 5 messages and 20 words in length
was classified, and the words used in conversations in each of these language collections 
were counted as in Section \ref{sec:curation}. (Two days is a short period
but plenty for the ranked most-frequent words from each language to stabilize.) Each of the top 100 words
for each language was checked to see if it was already in the classifier's word list for that language, and if not, it 
was added, ranked in the position it achieved in the word ranks for the conversations classified as belonging to this language.
Similar to Table \ref{table:english_words}, the top few words added for Chinese, Russian, French and German are shown in 
Table \ref{table:lp_words}, along with their corresponding ranks in the conversational and Wiki data sets. 

\begin{table}[t]
    \caption{Word-Ranks from Different Languages in Wikipedia and Conversational Data}
    \label{table:lp_words}

\begin{adjustbox}{width=\linewidth}

\renewcommand{\arraystretch}{1.5}
\centering
\begin{tabular}{c c}
\large Chinese & \large French \\
    \begin{tabular}{|p{1.8cm}|p{2.2cm}|r|r|}
    \hline
    `Word' & English & Conv & Wiki \\
    \hline
\begin{CJK*}{UTF8}{gbsn} 您正在與 \end{CJK*} & \raggedright You are working with  &  1  &  - \\
\begin{CJK*}{UTF8}{gbsn}  你好 \end{CJK*}  &  Hello there  &  2  &  250159 \\
\begin{CJK*}{UTF8}{gbsn}  會在  \end{CJK*} &  Will be at  &  4  & -  \\
\begin{CJK*}{UTF8}{gbsn}  字聯絡我們的客戶服務員  \end{CJK*} & \raggedright Contact our customer service staff  &  5  & -  \\
\begin{CJK*}{UTF8}{gbsn}  好的  \end{CJK*} &  Ok  &  6  & -  \\

\hline
    \end{tabular}
    &
    
\begin{tabular}{|l|l|r|r|}
    \hline
    Word & English & Conv & Wiki \\
    \hline
      votre  &  your  &  5  &  8116 \\
  bonjour  &  Hello  &  13  &  27605 \\
  merci  &  thank you  &  21  &  14647 \\
  pouvez  &  can  &  24  &  25874 \\
  bienvenue  &  welcome  &  37  &  22078 \\
  messaging  &  messaging  &  38  &  116417 \\
  avez  &  have  &  40  &  17100 \\
\hline
\end{tabular} 
 \\
 \\
\large Russian & \large German \\

\selectlanguage{russian}
\begin{tabular}{|l|l|r|r|}
    \hline
    Word & English & Conv & Wiki \\
\hline
      вам  &  you  &  1  &  5520\\
  пожалуйста &
  \makecell[l]{please / \\ you're welcome}  &  5  &  57985\\
  заказ  &  order  &  12  &  9541\\
  здравствуйте  &  Hello  &  14  &  285908\\
  могу  &  can  &  15  &  8265\\
  вас  &  you  &  18  &  5202\\
  выберите  &  select  &  20  &  798456\\
  \hline
\end{tabular} 

\selectlanguage{english}

&
\begin{tabular}{|l|l|r|r|}
    \hline
    Word & English & Conv & Wiki \\
    \hline
  bitte  &  \makecell[l]{please / \\ you're welcome}  &  5  &  7724 \\
  dich  &  you  &  23  &  7700 \\
  anliegen  &  issue  &  28  &  6405 \\
  https  &  https  &  33  &  52217 \\
  dir  &  to you  &  36  &  8899 \\
  hallo  &  Hi there  &  40  &  40001 \\
  danke  &  thanks  &  48  &  70621 \\
\hline
\end{tabular} 
\end{tabular} 

\end{adjustbox}

\end{table}

The classifier was then reloaded and rerun with these additional words.\footnote{In the software package, this corresponds to adding
the words from the file \texttt{data\_overrides.py}.}  Evaluating over 50K conversations and 1M messages,
we found that 1.69\% of the conversations and 7.9\% of the messages were classified differently. 
The latter number is larger than several of the differences in performance between the various classifiers measured in Section~\ref{sec:evaluation}. 

On an {\it ad hoc} sample of 100 conversations whose classification had changed, we found that 84 of the changes were improvements, 7 were 
worse, and the remaining 9 were neither.
Thanks to the simple and transparent nature of the classifier, we could see exactly which words had made the difference --- for example, the
phrase ``Bom dia'' was reclassified from Indonesian (where {\it dia} is a common third-person pronoun) to Portuguese (which is correct
in this case) by the boosting of the word {\it bom} (`good') from Portuguese conversational data.
Even though the Reciprocal Rank Classifier performed consistently well on evaluation tasks and was often the best performer,
there was still around a 7.9\% room for improvement on conversational messages! 

However, the whole
conversations that were reclassified in this step were much fewer and typically short (average 2.7 messages per conversation, 
compared to an average of above 19 for the whole dataset). The improvements demonstrated on these conversations did not translate
to improved evaluation results on the {\it Twituser} dataset in Section \ref{sec:twitter_results}, so this step appears to be less
useful in general that we originally thought.

\section{Multilingual Messages and Language Similarities}

Many individual words and messages could be understood in several languages. The single-word message ``No'' has already been mentioned,
and there are similar cases such as ``OK'' and ``Service''. Ideally, a language classifier should distinguish such cases: if two candidates
get the same score, it might be that the classifier reckons that there is one correct answer and each is equally likely; or it might be
that {\it both classifications are equally acceptable}. Statistical analysis of typical scoring ranges may enable us to 
distinguish such cases, and to predict explicitly that the message ``OK, baik'' (``OK, fine'') can be well-understood by
speakers of Indonesian {\it and} Malaysian.

It is sometimes possible to anticipate classification errors based on linguistic structure and related languages.
For example, the Serbian Wikipedia\footnote{\url{http//sr.wikipedia.org}} is predominantly written using the official 
Cyrillic alphabet, whereas less formal Serbian is often written using the Roman alphabet. It is highly predictable, from a knowledge of the history of the languages, that the \rrc will misclassify romanized Serbian as Croatian, which is always written using the Roman alphabet. This hypothesis was confirmed in a simple test: these two languages were added to the classifier, and then a page from
a popular radio website\footnote{\url{https://hitfm.rs/emisije/}} failed to score highly on the character test for Serbian and was
misclassified as Croatian instead.
At need, perhaps in response to feedback from Serbian speakers, we could mitigate the impact this error by adding suitable Serbian words to the word lists and increasing the relative frequency of Roman characters. This would not necessarily solve the problem, but it is an appropriate response to feedback. Of course, this could also be done for other similarly confusible languages.

In principle, the confusion patterns of the classifier could be used to organize the target languages into a hierarchical clustering. 
This would sometimes match linguistically expected similarities, and sometimes not. Japanese and Chinese are deemed similar, presumably because Chinese characters occur in written Japanese. But Korean is not added to this group, and the classifier does not see Arabic and Hebrew as similar at all. The reason is obvious: the classifier cannot see past the differences in character sets. Somewhat more subtly, when seen through the lens of averaged metrics, the numbers strongly depend on the exact choice of language and domain used for training and testing. Wikipedia, being an online encyclopedia, is formal, wide-ranging in vocabulary and particularly prone to interpolations of one language (often English) into texts that mainly use another language. Language politics also plays a part: it is easy to understand why contributors to the Serbian Wikipedia would be particularly keen to use aspects of the language, notably the Cyrillic writing system, that make it distinctive and special.

\section{The \texttt{lplangid} Package: Design and Usage}

The \rrc is available as an open source python package called \texttt{lplangid},\footnote{\scriptsize{\url{https://github.com/LivePersonInc/lplangid}}}
and can be installed from PyPi.\footnote{See \scriptsize{\url{https://pypi.org/project/lplangid/}} and install using \scriptsize{\texttt{pip install lplangid}}}
This section highlights parts of the package design that are of scientific and computational interest.

All of the classifiers tested in this paper are small compared to many contemporary NLP systems. For example, unzipped sizes on disk are 126MB for \fasttext 
with an optimized version at 917KB, 2.7MB for \langid, 32MB for SVC, 2.3MB for RRC. 
Sizes and speeds at runtime are similarly small for these classifiers. Thus, in systems that use a collection of NLP components, 
language classification is rarely a main concern for computational cost. The RRC can be easily optimized further by 
removing files for unwanted languages and words that are rarely encountered in practice, though such pruning is unlikely to be motivated
by physical cost.

The python code is designed to be small and flexible rather than full-featured. The main language classifier class is currently 212 lines of python code,
and the functions and constants described in Section \ref{sec:rrc} are easy to find and change. This is important because it enables flexibility in the
interface as well as the implementation. 

The input to language classifiers is nearly always a string, and the main design principle motivated by 
the experiments in this paper is to use long inputs where available. 
Desirable output formats are more varied, and include:

\begin{itemize}

\item \texttt{get\_winner => string}. Returns the 2-letter ISO-639-1 code of the single `winning' language. This is the most
commonly used interface to language classifiers, and implements the assumption ``there is a single correct language''.
However, the RRC package does not (currently) implement this.

\item \texttt{get\_winner => string or None}. The function in the \texttt{lplangid} package implements this version.
As above, but allowing for abstention. Within the use of abstention, there are 
different cases: abstaining on the message {\it saedfsadgrtfs} might mean ``this is unintelligible in any known language'', whereas abstaining
on the message {\it 123} might mean ``this is intelligible in all of these languages''. 
The \texttt{lplangid} package does not (currently) formally distinguish these cases.

\item \texttt{get\_winner\_score => (string, float)}. Includes the score of the winning language. This can be used as a cutoff / confidence threshold.
Distributions of scores vary for different lengths and genres of text, and appropriate thresholds should be determined on a case-by-case basis.

\item \texttt{get\_language\_scores => List[(string, float)]}. Includes the scores of all available languages. This 
can be used to pick the single best as above, and for other purposes. Comparing the best score with the runners-up can be used
as a confidence measure. Sometimes one of the runners-up may be preferred if it is better-supported.

\end{itemize}

Each of these patterns has its uses. As an example, a component that performed named entity recognition had models for English and Spanish.
As suggested above, short messages in Spanish are sometimes confused with Italian. These cases were processed more effectively with the logic
``get a list of languages and scores, and if there is no model for the best-scoring language, try the next best'', 
rather than the logic ``get the best-scoring language, and if there is no model for that, give up''. 
When adding new languages to the classifier, such consequences should be considered: if a new language (say Malaysian)
is introduced, it is may take results from similar languages (in this case Indonesian). If Malaysian is not
supported as well by other system components and there is no fallback logic to say ``if Malaysian doesn't work then try Indonesian'', 
the addition of new languages can improve the language classifier component but harm the system overall.

In our experience, undesirable outcomes from language classifiers are often because these case scenarios have not been thought through 
as part of the design process. We recommend the RRC from the \texttt{lplangid} package partly because it is easy to adjust its languages and
interfaces. More importantly, we believe that in most situations today, several language classifiers can be found that report good results
on several tasks --- but system designers should not assume that language classifiers are all more or less the same. It is vital to
understand how the language classifier outputs are used throughout the rest of the system to make sure that the overall behavior
is correct, and that the classifier's behavior meets the needs of the other subsystems. 

\section{Conclusions}

The reciprocal rank of a feature in a frequency list can be used in a scoring function for a simple classifier.
The main machine learning conclusion of this paper is that a \rrc is simple and competitive for language detection, 
especially in combination with out-of-the-box \fasttext. Though trained on Wikipedia, 
the RRC performs well on Twitter data --- the classifier shows robustness to this domain transfer,
and performance on commercially significant data suggests that this conclusion can be generalized.

The main system engineering conclusion of this paper is that \rrc is highly maintainable and benefits from straightforward 
approaches to data curation. It is easier to understand and edit a list of frequent words than it is to adjust a 
training set and/or training algorithm, let alone editing other classifier model files directly.
Changes to the RRC are quick, cheap, and easy to verify with unit tests.
Simple linguistic insights about the typology, 
context and use of particular languages are easy to incorporate, evaluate and extend. 
We can still benefit from improvements made by the wider machine learning community, 
since simple ensembles combining a state of the art machine-learned classifier with \rrc sometimes 
perform better than either component alone.

It is common today to cast language processing in general as a collection of supervised or unsupervised machine learning problems, 
and to assume
that the main advances will come from more data and more sophisticated ML models. One final conclusion is that this is not
always the case: the design and experimental success of the \rrc demonstrates that if we start with the question
``What do we know about {\it languages}?'', then sometimes a simpler, more versatile, and explicit solution
may still be found.

\bibliography{anthology,acl2020,semanticvectors}
\bibliographystyle{acl_natbib}

\end{document}

\section{Full tables : Wikipedia}

\subsection{SVC}

\begin{table}[h]
    \centering
\begin{tabular}{lrrrr}
\toprule
{} &  precision &  recall &  f1-score &  support \\
\midrule
ar           &      99.43 &   98.86 &     99.14 &      351 \\
de           &      94.82 &   94.18 &     94.50 &      447 \\
el           &      99.64 &   96.55 &     98.07 &      290 \\
en           &      79.91 &   97.02 &     87.64 &      369 \\
es           &      93.38 &   95.01 &     94.19 &      401 \\
fr           &      88.03 &   95.24 &     91.49 &      525 \\
he           &     100.00 &   99.73 &     99.87 &      376 \\
hi           &      99.55 &   99.55 &     99.55 &      221 \\
id           &      96.90 &   80.95 &     88.21 &      656 \\
it           &      93.24 &   84.66 &     88.75 &      652 \\
ja           &      99.54 &   91.56 &     95.38 &      474 \\
ko           &      99.84 &   97.09 &     98.44 &      652 \\
mk           &      99.69 &   88.80 &     93.93 &      366 \\
nl           &      95.88 &   75.56 &     84.51 &      585 \\
pt           &      97.42 &   92.45 &     94.87 &      490 \\
ru           &      99.63 &   95.68 &     97.61 &      278 \\
sl           &      97.23 &   84.59 &     90.47 &      623 \\
sq           &      98.90 &   81.60 &     89.42 &      549 \\
th           &     100.00 &   97.38 &     98.67 &      267 \\
tl           &      91.07 &   72.60 &     80.79 &      365 \\
vi           &      96.83 &   80.00 &     87.62 &      535 \\
zh           &      98.65 &   96.97 &     97.80 &      528 \\
accuracy     &      89.58 &   89.58 &     89.58 &        0 \\
macro avg    &      96.17 &   90.55 &     93.05 &    10000 \\
weighted avg &      96.23 &   89.58 &     92.55 &    10000 \\
\bottomrule
\end{tabular}

    \caption{RRC on Wikipedia samples of size 256}
    \label{tab:rrc256}
\end{table}

\begin{table}[h]
    \centering
    \begin{tabular}{lrrrr}
\toprule
{} &  precision &  recall &  f1-score &  support \\
\midrule
ar           &      99.42 &   98.01 &     98.71 &      351 \\
de           &      89.51 &   82.10 &     85.65 &      447 \\
el           &      98.57 &   95.17 &     96.84 &      290 \\
en           &      68.54 &   89.16 &     77.50 &      369 \\
es           &      80.51 &   78.30 &     79.39 &      401 \\
fr           &      80.11 &   86.67 &     83.26 &      525 \\
he           &     100.00 &   98.40 &     99.20 &      376 \\
hi           &      99.55 &   99.10 &     99.32 &      221 \\
id           &      93.93 &   73.17 &     82.26 &      656 \\
it           &      88.57 &   74.85 &     81.13 &      652 \\
ja           &      99.51 &   85.02 &     91.70 &      474 \\
ko           &      99.68 &   95.71 &     97.65 &      652 \\
mk           &      96.81 &   82.79 &     89.25 &      366 \\
nl           &      88.58 &   66.32 &     75.86 &      585 \\
pt           &      90.84 &   76.94 &     83.31 &      490 \\
ru           &      98.98 &   69.78 &     81.86 &      278 \\
sl           &      93.00 &   72.55 &     81.51 &      623 \\
sq           &      98.05 &   73.22 &     83.84 &      549 \\
th           &     100.00 &   93.26 &     96.51 &      267 \\
tl           &      83.50 &   69.32 &     75.75 &      365 \\
vi           &      95.09 &   76.07 &     84.53 &      535 \\
zh           &      98.19 &   92.42 &     95.22 &      528 \\
accuracy     &      81.82 &   81.82 &     81.82 &        0 \\
macro avg    &      92.60 &   82.95 &     87.12 &    10000 \\
weighted avg &      92.52 &   81.82 &     86.47 &    10000 \\
\bottomrule
\end{tabular}

    \caption{RRC on Wikipedia samples of size 16}
    \label{tab:rrc16}
\end{table}

\begin{table}[h]
    \centering
    \begin{tabular}{lrrrr}
\toprule
{} &  precision &  recall &  f1-score &  support \\
\midrule
ar           &      98.85 &   97.72 &     98.28 &      351 \\
de           &      92.52 &   83.00 &     87.50 &      447 \\
el           &      99.27 &   93.79 &     96.45 &      290 \\
en           &      85.80 &   80.22 &     82.91 &      369 \\
es           &      88.30 &   79.05 &     83.42 &      401 \\
fr           &      86.61 &   81.33 &     83.89 &      525 \\
he           &     100.00 &   98.14 &     99.06 &      376 \\
hi           &      99.54 &   97.74 &     98.63 &      221 \\
id           &      68.93 &   77.44 &     72.94 &      656 \\
it           &      94.32 &   73.93 &     82.89 &      652 \\
ja           &      99.30 &   89.24 &     94.00 &      474 \\
ko           &      98.64 &   88.80 &     93.46 &      652 \\
mk           &      96.87 &   84.43 &     90.22 &      366 \\
nl           &      37.21 &   83.08 &     51.40 &      585 \\
pt           &      86.06 &   79.39 &     82.59 &      490 \\
ru           &      88.93 &   92.45 &     90.65 &      278 \\
sl           &      79.31 &   81.22 &     80.25 &      623 \\
sq           &      98.27 &   72.50 &     83.44 &      549 \\
th           &     100.00 &   93.26 &     96.51 &      267 \\
tl           &      84.10 &   65.21 &     73.46 &      365 \\
vi           &      83.06 &   77.01 &     79.92 &      535 \\
zh           &      96.38 &   90.72 &     93.46 &      528 \\
accuracy     &      83.26 &   83.26 &     83.26 &        0 \\
macro avg    &      89.19 &   84.53 &     86.15 &    10000 \\
weighted avg &      87.53 &   83.26 &     84.54 &    10000 \\
\bottomrule
\end{tabular}

    \caption{RRC on Wikipedia samples of size 32}
    \label{tab:rrc32}
\end{table}

\begin{table}[h]
    \centering
    \begin{tabular}{lrrrr}
\toprule
{} &  precision &  recall &  f1-score &  support \\
\midrule
ar           &      99.14 &   98.58 &     98.86 &      351 \\
de           &      95.93 &   89.71 &     92.72 &      447 \\
el           &      98.58 &   95.52 &     97.02 &      290 \\
en           &      87.16 &   86.45 &     86.80 &      369 \\
es           &      90.72 &   87.78 &     89.23 &      401 \\
fr           &      91.73 &   88.76 &     90.22 &      525 \\
he           &     100.00 &   98.94 &     99.47 &      376 \\
hi           &      99.54 &   97.74 &     98.63 &      221 \\
id           &      75.11 &   79.57 &     77.28 &      656 \\
it           &      95.34 &   78.37 &     86.03 &      652 \\
ja           &      99.53 &   89.45 &     94.22 &      474 \\
ko           &      98.33 &   90.18 &     94.08 &      652 \\
mk           &      96.32 &   85.79 &     90.75 &      366 \\
nl           &      41.19 &   85.47 &     55.59 &      585 \\
pt           &      91.89 &   85.51 &     88.58 &      490 \\
ru           &      90.53 &   92.81 &     91.65 &      278 \\
sl           &      83.88 &   82.66 &     83.27 &      623 \\
sq           &      98.82 &   76.14 &     86.01 &      549 \\
th           &     100.00 &   95.13 &     97.50 &      267 \\
tl           &      89.68 &   69.04 &     78.02 &      365 \\
vi           &      85.83 &   79.25 &     82.41 &      535 \\
zh           &      96.99 &   91.67 &     94.26 &      528 \\
accuracy     &      86.32 &   86.32 &     86.32 &        0 \\
macro avg    &      91.19 &   87.48 &     88.75 &    10000 \\
weighted avg &      89.79 &   86.32 &     87.35 &    10000 \\
\bottomrule
\end{tabular}

    \caption{RRC on Wikipedia samples of size 64}
    \label{tab:rrc64}
\end{table}

\subsection{SVC}
\begin{table}[h]
    \centering
    \begin{tabular}{lrrrr}
\toprule
{} &  precision &  recall &  f1-score &  support \\
\midrule
ar           &      98.86 &   98.58 &     98.72 &      351 \\
de           &      96.97 &   93.06 &     94.98 &      447 \\
el           &      99.29 &   96.21 &     97.72 &      290 \\
en           &      90.98 &   92.95 &     91.96 &      369 \\
es           &      93.48 &   93.02 &     93.25 &      401 \\
fr           &      96.47 &   93.71 &     95.07 &      525 \\
he           &     100.00 &   99.47 &     99.73 &      376 \\
hi           &      99.54 &   98.64 &     99.09 &      221 \\
id           &      78.62 &   81.86 &     80.21 &      656 \\
it           &      96.27 &   83.13 &     89.22 &      652 \\
ja           &      99.54 &   90.51 &     94.81 &      474 \\
ko           &      98.49 &   90.18 &     94.16 &      652 \\
mk           &      98.51 &   90.16 &     94.15 &      366 \\
nl           &      44.09 &   84.10 &     57.85 &      585 \\
pt           &      96.10 &   90.61 &     93.28 &      490 \\
ru           &      95.37 &   96.40 &     95.89 &      278 \\
sl           &      86.99 &   85.87 &     86.43 &      623 \\
sq           &      99.31 &   78.51 &     87.69 &      549 \\
th           &     100.00 &   96.63 &     98.29 &      267 \\
tl           &      87.88 &   71.51 &     78.85 &      365 \\
vi           &      87.98 &   80.75 &     84.21 &      535 \\
zh           &      97.79 &   92.23 &     94.93 &      528 \\
accuracy     &      88.75 &   88.75 &     88.75 &        0 \\
macro avg    &      92.84 &   89.91 &     90.93 &    10000 \\
weighted avg &      91.55 &   88.75 &     89.60 &    10000 \\
\bottomrule
\end{tabular}

    \caption{SVC on Wikipedia samples of size 256}
    \label{tab:svc256}
\end{table}

\begin{table}[h]
    \centering
    \begin{tabular}{lrrrr}
\toprule
{} &  precision &  recall &  f1-score &  support \\
\midrule
ar           &      99.13 &   97.15 &     98.13 &      351 \\
de           &      91.48 &   64.88 &     75.92 &      447 \\
el           &      98.92 &   94.48 &     96.65 &      290 \\
en           &      81.74 &   53.39 &     64.59 &      369 \\
es           &      76.83 &   60.35 &     67.60 &      401 \\
fr           &      85.41 &   68.00 &     75.72 &      525 \\
he           &     100.00 &   97.07 &     98.52 &      376 \\
hi           &      99.53 &   96.38 &     97.93 &      221 \\
id           &      60.59 &   71.49 &     65.59 &      656 \\
it           &      91.28 &   57.82 &     70.80 &      652 \\
ja           &      98.57 &   87.34 &     92.62 &      474 \\
ko           &      98.44 &   87.27 &     92.52 &      652 \\
mk           &      92.53 &   77.87 &     84.57 &      366 \\
nl           &      25.76 &   80.00 &     38.97 &      585 \\
pt           &      80.37 &   61.84 &     69.90 &      490 \\
ru           &      81.88 &   87.77 &     84.72 &      278 \\
sl           &      70.23 &   68.54 &     69.37 &      623 \\
sq           &      97.90 &   59.38 &     73.92 &      549 \\
th           &     100.00 &   91.01 &     95.29 &      267 \\
tl           &      78.55 &   62.19 &     69.42 &      365 \\
vi           &      70.68 &   73.46 &     72.04 &      535 \\
zh           &      95.56 &   89.58 &     92.47 &      528 \\
accuracy     &      74.97 &   74.97 &     74.97 &        0 \\
macro avg    &      85.24 &   76.69 &     79.42 &    10000 \\
weighted avg &      83.22 &   74.97 &     77.27 &    10000 \\
\bottomrule
\end{tabular}

    \caption{SVC on Wikipedia samples of size 16}
    \label{tab:svc16}
\end{table}

\begin{table}[h]
    \centering
    \begin{tabular}{lrrrr}
\toprule
{} &  precision &  recall &  f1-score &  support \\
\midrule
ar           &      98.85 &   97.72 &     98.28 &      351 \\
de           &      92.52 &   83.00 &     87.50 &      447 \\
el           &      99.27 &   93.79 &     96.45 &      290 \\
en           &      85.80 &   80.22 &     82.91 &      369 \\
es           &      88.30 &   79.05 &     83.42 &      401 \\
fr           &      86.61 &   81.33 &     83.89 &      525 \\
he           &     100.00 &   98.14 &     99.06 &      376 \\
hi           &      99.54 &   97.74 &     98.63 &      221 \\
id           &      68.93 &   77.44 &     72.94 &      656 \\
it           &      94.32 &   73.93 &     82.89 &      652 \\
ja           &      99.30 &   89.24 &     94.00 &      474 \\
ko           &      98.64 &   88.80 &     93.46 &      652 \\
mk           &      96.87 &   84.43 &     90.22 &      366 \\
nl           &      37.21 &   83.08 &     51.40 &      585 \\
pt           &      86.06 &   79.39 &     82.59 &      490 \\
ru           &      88.93 &   92.45 &     90.65 &      278 \\
sl           &      79.31 &   81.22 &     80.25 &      623 \\
sq           &      98.27 &   72.50 &     83.44 &      549 \\
th           &     100.00 &   93.26 &     96.51 &      267 \\
tl           &      84.10 &   65.21 &     73.46 &      365 \\
vi           &      83.06 &   77.01 &     79.92 &      535 \\
zh           &      96.38 &   90.72 &     93.46 &      528 \\
accuracy     &      83.26 &   83.26 &     83.26 &        0 \\
macro avg    &      89.19 &   84.53 &     86.15 &    10000 \\
weighted avg &      87.53 &   83.26 &     84.54 &    10000 \\
\bottomrule
\end{tabular}

    \caption{SVC on Wikipedia samples of size 32}
    \label{tab:svc32}
\end{table}

\begin{table}[h]
    \centering
    \begin{tabular}{lrrrr}
\toprule
{} &  precision &  recall &  f1-score &  support \\
\midrule
ar           &      99.14 &   98.58 &     98.86 &      351 \\
de           &      95.93 &   89.71 &     92.72 &      447 \\
el           &      98.58 &   95.52 &     97.02 &      290 \\
en           &      87.16 &   86.45 &     86.80 &      369 \\
es           &      90.72 &   87.78 &     89.23 &      401 \\
fr           &      91.73 &   88.76 &     90.22 &      525 \\
he           &     100.00 &   98.94 &     99.47 &      376 \\
hi           &      99.54 &   97.74 &     98.63 &      221 \\
id           &      75.11 &   79.57 &     77.28 &      656 \\
it           &      95.34 &   78.37 &     86.03 &      652 \\
ja           &      99.53 &   89.45 &     94.22 &      474 \\
ko           &      98.33 &   90.18 &     94.08 &      652 \\
mk           &      96.32 &   85.79 &     90.75 &      366 \\
nl           &      41.19 &   85.47 &     55.59 &      585 \\
pt           &      91.89 &   85.51 &     88.58 &      490 \\
ru           &      90.53 &   92.81 &     91.65 &      278 \\
sl           &      83.88 &   82.66 &     83.27 &      623 \\
sq           &      98.82 &   76.14 &     86.01 &      549 \\
th           &     100.00 &   95.13 &     97.50 &      267 \\
tl           &      89.68 &   69.04 &     78.02 &      365 \\
vi           &      85.83 &   79.25 &     82.41 &      535 \\
zh           &      96.99 &   91.67 &     94.26 &      528 \\
accuracy     &      86.32 &   86.32 &     86.32 &        0 \\
macro avg    &      91.19 &   87.48 &     88.75 &    10000 \\
weighted avg &      89.79 &   86.32 &     87.35 &    10000 \\
\bottomrule
\end{tabular}

    \caption{SVC on Wikipedia samples of size 64}
    \label{tab:svc64}
\end{table}

\subsection{RRC + Fasttext}
\begin{table}[h]
    \centering
    \begin{tabular}{lrrrr}
\toprule
{} &  precision &  recall &  f1-score &  support \\
\midrule
ar           &      96.93 &   98.86 &     97.88 &      351 \\
de           &      86.17 &   96.20 &     90.91 &      447 \\
el           &      99.29 &   96.55 &     97.90 &      290 \\
en           &      52.92 &   98.10 &     68.76 &      369 \\
es           &      86.13 &   96.01 &     90.80 &      401 \\
fr           &      79.25 &   96.00 &     86.82 &      525 \\
he           &      99.47 &   99.73 &     99.60 &      376 \\
hi           &      99.55 &   99.55 &     99.55 &      221 \\
id           &      95.40 &   82.16 &     88.29 &      656 \\
it           &      86.59 &   86.20 &     86.40 &      652 \\
ja           &      97.47 &   97.68 &     97.58 &      474 \\
ko           &      99.69 &   97.55 &     98.60 &      652 \\
mk           &      99.71 &   94.54 &     97.05 &      366 \\
nl           &      93.66 &   80.85 &     86.79 &      585 \\
pt           &      93.67 &   93.67 &     93.67 &      490 \\
ru           &      93.20 &   98.56 &     95.80 &      278 \\
sl           &      96.56 &   85.71 &     90.82 &      623 \\
sq           &      98.47 &   82.15 &     89.57 &      549 \\
th           &     100.00 &   97.38 &     98.67 &      267 \\
tl           &      91.07 &   72.60 &     80.79 &      365 \\
vi           &      96.62 &   80.19 &     87.64 &      535 \\
zh           &      98.10 &   97.92 &     98.01 &      528 \\
accuracy     &      91.11 &   91.11 &     91.11 &        0 \\
macro avg    &      92.72 &   92.19 &     91.91 &    10000 \\
weighted avg &      92.74 &   91.11 &     91.40 &    10000 \\
\bottomrule
\end{tabular}

    \caption{RRC + FastText on Wikipedia samples of size 256}
    \label{tab:hybrid256}
\end{table}

\begin{table}[h]
    \centering
    \begin{tabular}{lrrrr}
\toprule
{} &  precision &   recall &  f1-score &  support \\
\midrule
ar           &    95.8217 &  98.0057 &   96.9014 &      351 \\
de           &    78.6127 &  91.2752 &   84.4720 &      447 \\
el           &    98.5714 &  95.1724 &   96.8421 &      290 \\
en           &    41.8227 &  90.7859 &   57.2650 &      369 \\
es           &    72.2101 &  82.2943 &   76.9231 &      401 \\
fr           &    70.1780 &  90.0952 &   78.8991 &      525 \\
he           &    99.4624 &  98.4043 &   98.9305 &      376 \\
hi           &    99.0991 &  99.5475 &   99.3228 &      221 \\
id           &    92.1495 &  75.1524 &   82.7876 &      656 \\
it           &    81.7035 &  79.4479 &   80.5599 &      652 \\
ja           &    97.2043 &  95.3586 &   96.2726 &      474 \\
ko           &    99.5268 &  96.7791 &   98.1337 &      652 \\
mk           &    97.1098 &  91.8033 &   94.3820 &      366 \\
nl           &    85.3175 &  73.5043 &   78.9715 &      585 \\
pt           &    86.0927 &  79.5918 &   82.7147 &      490 \\
ru           &    89.6552 &  93.5252 &   91.5493 &      278 \\
sl           &    92.6641 &  77.0465 &   84.1367 &      623 \\
sq           &    97.6190 &  74.6812 &   84.6233 &      549 \\
th           &   100.0000 &  93.2584 &   96.5116 &      267 \\
tl           &    83.2787 &  69.5890 &   75.8209 &      365 \\
vi           &    94.6636 &  76.2617 &   84.4720 &      535 \\
zh           &    95.8647 &  96.5909 &   96.2264 &      528 \\
accuracy     &    85.7700 &  85.7700 &   85.7700 &        0 \\
macro avg    &    88.5740 &  87.1896 &   87.1236 &    10000 \\
weighted avg &    88.4820 &  85.7700 &   86.3895 &    10000 \\
\bottomrule
\end{tabular}

    \caption{RRC + FastText on Wikipedia samples of size 16}
    \label{tab:hybrid16}
\end{table}

\begin{table}[h]
    \centering
    \begin{tabular}{lrrrr}
\toprule
{} &  precision &  recall &  f1-score &  support \\
\midrule
ar           &      96.09 &   98.01 &     97.04 &      351 \\
de           &      82.72 &   95.30 &     88.57 &      447 \\
el           &     100.00 &   93.79 &     96.80 &      290 \\
en           &      47.72 &   96.48 &     63.86 &      369 \\
es           &      81.35 &   90.27 &     85.58 &      401 \\
fr           &      75.12 &   93.14 &     83.16 &      525 \\
he           &      99.73 &   99.47 &     99.60 &      376 \\
hi           &      99.10 &   99.10 &     99.10 &      221 \\
id           &      93.90 &   79.73 &     86.23 &      656 \\
it           &      84.36 &   81.90 &     83.11 &      652 \\
ja           &      97.25 &   97.05 &     97.15 &      474 \\
ko           &      99.69 &   97.09 &     98.37 &      652 \\
mk           &      99.42 &   93.17 &     96.19 &      366 \\
nl           &      90.91 &   76.92 &     83.33 &      585 \\
pt           &      89.80 &   89.80 &     89.80 &      490 \\
ru           &      91.22 &   97.12 &     94.08 &      278 \\
sl           &      95.59 &   83.47 &     89.12 &      623 \\
sq           &      98.20 &   79.42 &     87.81 &      549 \\
th           &     100.00 &   96.63 &     98.29 &      267 \\
tl           &      89.75 &   69.59 &     78.40 &      365 \\
vi           &      96.10 &   78.32 &     86.30 &      535 \\
zh           &      96.79 &   97.16 &     96.98 &      528 \\
accuracy     &      88.93 &   88.93 &     88.93 &        0 \\
macro avg    &      91.13 &   90.13 &     89.95 &    10000 \\
weighted avg &      91.08 &   88.93 &     89.36 &    10000 \\
\bottomrule
\end{tabular}

    \caption{RRC + FastText on Wikipedia samples of size 32}
    \label{tab:hybrid32}
\end{table}

\begin{table}[h]
    \centering
    \begin{tabular}{lrrrr}
\toprule
{} &  precision &  recall &  f1-score &  support \\
\midrule
ar           &      96.66 &   98.86 &     97.75 &      351 \\
de           &      83.86 &   95.30 &     89.21 &      447 \\
el           &      98.58 &   95.86 &     97.20 &      290 \\
en           &      49.72 &   97.56 &     65.87 &      369 \\
es           &      83.11 &   93.27 &     87.90 &      401 \\
fr           &      78.41 &   94.10 &     85.54 &      525 \\
he           &      99.73 &   99.73 &     99.73 &      376 \\
hi           &      99.10 &   99.55 &     99.32 &      221 \\
id           &      94.45 &   80.49 &     86.91 &      656 \\
it           &      84.79 &   84.66 &     84.73 &      652 \\
ja           &      97.87 &   96.84 &     97.35 &      474 \\
ko           &      99.69 &   97.39 &     98.53 &      652 \\
mk           &      97.69 &   92.35 &     94.94 &      366 \\
nl           &      92.28 &   77.61 &     84.31 &      585 \\
pt           &      92.80 &   92.04 &     92.42 &      490 \\
ru           &      90.17 &   95.68 &     92.84 &      278 \\
sl           &      96.13 &   83.63 &     89.44 &      623 \\
sq           &      98.44 &   80.33 &     88.47 &      549 \\
th           &     100.00 &   97.38 &     98.67 &      267 \\
tl           &      91.26 &   71.51 &     80.18 &      365 \\
vi           &      96.38 &   79.63 &     87.21 &      535 \\
zh           &      96.64 &   98.11 &     97.37 &      528 \\
accuracy     &      89.84 &   89.84 &     89.84 &        0 \\
macro avg    &      91.72 &   90.99 &     90.72 &    10000 \\
weighted avg &      91.77 &   89.84 &     90.21 &    10000 \\
\bottomrule
\end{tabular}

    \caption{RRC + FastText on Wikipedia samples of size 64}
    \label{tab:hybrid64}
\end{table}

\subsection{FastText}
\begin{table}[h]
    \centering
    \begin{tabular}{lrrrr}
\toprule
{} &  precision &   recall &  f1-score &  support \\
\midrule
ar           &    95.3297 &  98.8604 &   97.0629 &      351 \\
de           &    84.5098 &  96.4206 &   90.0731 &      447 \\
el           &    99.6441 &  96.5517 &   98.0736 &      290 \\
en           &    43.3850 &  98.6450 &   60.2649 &      369 \\
es           &    85.7143 &  97.2569 &   91.1215 &      401 \\
fr           &    81.7886 &  95.8095 &   88.2456 &      525 \\
he           &    99.4695 &  99.7340 &   99.6016 &      376 \\
hi           &    99.0950 &  99.0950 &   99.0950 &      221 \\
id           &    97.1805 &  78.8110 &   87.0370 &      656 \\
it           &    88.4555 &  86.9632 &   87.7030 &      652 \\
ja           &    95.5102 &  98.7342 &   97.0954 &      474 \\
ko           &    99.6835 &  96.6258 &   98.1308 &      652 \\
mk           &    99.7059 &  92.6230 &   96.0340 &      366 \\
nl           &    95.8932 &  79.8291 &   87.1269 &      585 \\
pt           &    93.8272 &  93.0612 &   93.4426 &      490 \\
ru           &    91.0596 &  98.9209 &   94.8276 &      278 \\
sl           &    99.2439 &  84.2697 &   91.1458 &      623 \\
sq           &    99.1011 &  80.3279 &   88.7324 &      549 \\
th           &    99.6169 &  97.3783 &   98.4848 &      267 \\
tl           &    99.5951 &  67.3973 &   80.3922 &      365 \\
vi           &    99.7658 &  79.6262 &   88.5655 &      535 \\
zh           &    98.4586 &  96.7803 &   97.6122 &      528 \\
accuracy     &    90.3700 &  90.3700 &   90.3700 &        0 \\
macro avg    &    93.0015 &  91.5328 &   91.3577 &    10000 \\
weighted avg &    93.2868 &  90.3700 &   90.9838 &    10000 \\
\bottomrule
\end{tabular}

    \caption{FastText on Wikipedia samples of size 256}
    \label{tab:fasttext256}
\end{table}

\begin{table}[h]
    \centering
    \begin{tabular}{lrrrr}
\toprule
{} &  precision &   recall &  f1-score &  support \\
\midrule
ar           &    93.4426 &  97.4359 &   95.3975 &      351 \\
de           &    74.6835 &  92.3937 &   82.6000 &      447 \\
el           &    99.6350 &  94.1379 &   96.8085 &      290 \\
en           &    31.4086 &  97.2900 &   47.4868 &      369 \\
es           &    67.8846 &  88.0299 &   76.6558 &      401 \\
fr           &    69.5394 &  89.1429 &   78.1302 &      525 \\
he           &    99.4624 &  98.4043 &   98.9305 &      376 \\
hi           &    99.0783 &  97.2851 &   98.1735 &      221 \\
id           &    93.1034 &  65.8537 &   77.1429 &      656 \\
it           &    79.5349 &  78.6810 &   79.1056 &      652 \\
ja           &    94.2387 &  96.6245 &   95.4167 &      474 \\
ko           &    99.5215 &  95.7055 &   97.5762 &      652 \\
mk           &    99.3691 &  86.0656 &   92.2401 &      366 \\
nl           &    89.9543 &  67.3504 &   77.0283 &      585 \\
pt           &    86.4055 &  76.5306 &   81.1688 &      490 \\
ru           &    83.2298 &  96.4029 &   89.3333 &      278 \\
sl           &    96.6587 &  65.0080 &   77.7351 &      623 \\
sq           &    98.6264 &  65.3916 &   78.6418 &      549 \\
th           &   100.0000 &  93.2584 &   96.5116 &      267 \\
tl           &    97.2826 &  49.0411 &   65.2095 &      365 \\
vi           &    98.7562 &  74.2056 &   84.7385 &      535 \\
zh           &    94.7269 &  95.2652 &   94.9953 &      528 \\
accuracy     &    82.6400 &  82.6400 &   82.6400 &        0 \\
macro avg    &    88.4792 &  84.5229 &   84.5921 &    10000 \\
weighted avg &    88.6071 &  82.6400 &   83.7780 &    10000 \\
\bottomrule
\end{tabular}

    \caption{FastText on Wikipedia samples of size 16}
    \label{tab:fasttext16}
\end{table}

\begin{table}[h]
    \centering
    \begin{tabular}{lrrrr}
\toprule
{} &  precision &  recall &  f1-score &  support \\
\midrule
ar           &      94.23 &   97.72 &     95.94 &      351 \\
de           &      79.59 &   95.97 &     87.02 &      447 \\
el           &     100.00 &   93.79 &     96.80 &      290 \\
en           &      37.56 &   98.64 &     54.41 &      369 \\
es           &      76.51 &   95.01 &     84.76 &      401 \\
fr           &      78.12 &   93.14 &     84.97 &      525 \\
he           &      99.73 &   99.47 &     99.60 &      376 \\
hi           &      98.64 &   98.64 &     98.64 &      221 \\
id           &      95.87 &   74.39 &     83.78 &      656 \\
it           &      84.66 &   82.98 &     83.81 &      652 \\
ja           &      95.49 &   98.31 &     96.88 &      474 \\
ko           &      99.68 &   96.32 &     97.97 &      652 \\
mk           &     100.00 &   88.52 &     93.91 &      366 \\
nl           &      94.59 &   74.70 &     83.48 &      585 \\
pt           &      91.68 &   85.51 &     88.49 &      490 \\
ru           &      86.39 &   98.20 &     91.92 &      278 \\
sl           &      98.15 &   76.57 &     86.02 &      623 \\
sq           &      99.04 &   75.05 &     85.39 &      549 \\
th           &     100.00 &   97.00 &     98.48 &      267 \\
tl           &      98.56 &   56.16 &     71.55 &      365 \\
vi           &      99.29 &   77.94 &     87.33 &      535 \\
zh           &      96.74 &   95.64 &     96.19 &      528 \\
accuracy     &      87.21 &   87.21 &     87.21 &        0 \\
macro avg    &      91.12 &   88.62 &     88.52 &    10000 \\
weighted avg &      91.41 &   87.21 &     88.05 &    10000 \\
\bottomrule
\end{tabular}

    \caption{FastText on Wikipedia samples of size 32}
    \label{tab:fasttext32}
\end{table}

\begin{table}[h]
    \centering
    \begin{tabular}{lrrrr}
\toprule
{} &  precision &  recall &  f1-score &  support \\
\midrule
ar           &      94.81 &   98.86 &     96.79 &      351 \\
de           &      82.15 &   95.75 &     88.43 &      447 \\
el           &      99.29 &   95.86 &     97.54 &      290 \\
en           &      40.38 &   98.92 &     57.34 &      369 \\
es           &      80.72 &   95.01 &     87.29 &      401 \\
fr           &      80.52 &   94.48 &     86.94 &      525 \\
he           &      99.73 &   99.73 &     99.73 &      376 \\
hi           &      98.64 &   98.19 &     98.41 &      221 \\
id           &      96.72 &   76.52 &     85.45 &      656 \\
it           &      87.46 &   84.51 &     85.96 &      652 \\
ja           &      95.87 &   97.89 &     96.87 &      474 \\
ko           &      99.53 &   96.78 &     98.13 &      652 \\
mk           &      99.39 &   89.62 &     94.25 &      366 \\
nl           &      95.30 &   76.24 &     84.71 &      585 \\
pt           &      92.07 &   90.00 &     91.02 &      490 \\
ru           &      87.78 &   98.20 &     92.70 &      278 \\
sl           &      99.21 &   80.42 &     88.83 &      623 \\
sq           &      99.08 &   78.51 &     87.60 &      549 \\
th           &     100.00 &   97.38 &     98.67 &      267 \\
tl           &      99.15 &   63.56 &     77.46 &      365 \\
vi           &      99.30 &   79.07 &     88.03 &      535 \\
zh           &      96.23 &   96.59 &     96.41 &      528 \\
accuracy     &      88.80 &   88.80 &     88.80 &        0 \\
macro avg    &      91.97 &   90.09 &     89.94 &    10000 \\
weighted avg &      92.31 &   88.80 &     89.53 &    10000 \\
\bottomrule
\end{tabular}

    \caption{FastText on Wikipedia samples of size 64}
    \label{tab:fasttext64}
\end{table}

\subsection{\texttt{langid.py}}
\begin{table}[h]
    \centering
    \begin{tabular}{lrrrr}
\toprule
{} &  precision &   recall &  f1-score &  support \\
\midrule
ar           &    98.8604 &  98.8604 &   98.8604 &      351 \\
de           &    79.5497 &  94.8546 &   86.5306 &      447 \\
el           &    99.2908 &  96.5517 &   97.9021 &      290 \\
en           &    40.9556 &  97.5610 &   57.6923 &      369 \\
es           &    80.2105 &  95.0125 &   86.9863 &      401 \\
fr           &    81.4935 &  95.6190 &   87.9930 &      525 \\
he           &    99.4695 &  99.7340 &   99.6016 &      376 \\
hi           &    99.5475 &  99.5475 &   99.5475 &      221 \\
id           &    96.1612 &  76.3720 &   85.1317 &      656 \\
it           &    86.8878 &  84.3558 &   85.6031 &      652 \\
ja           &    99.0991 &  92.8270 &   95.8606 &      474 \\
ko           &    99.8435 &  97.8528 &   98.8381 &      652 \\
mk           &    99.3808 &  87.7049 &   93.1785 &      366 \\
nl           &    96.2555 &  74.7009 &   84.1193 &      585 \\
pt           &    95.3191 &  91.4286 &   93.3333 &      490 \\
ru           &    88.0645 &  98.2014 &   92.8571 &      278 \\
sl           &    96.9868 &  82.6645 &   89.2548 &      623 \\
sq           &    99.0950 &  79.7814 &   88.3956 &      549 \\
th           &   100.0000 &  97.3783 &   98.6717 &      267 \\
tl           &    94.5525 &  66.5753 &   78.1350 &      365 \\
vi           &    99.5338 &  79.8131 &   88.5892 &      535 \\
zh           &    93.8517 &  98.2955 &   96.0222 &      528 \\
accuracy     &    88.9900 &  88.9900 &   88.9900 &        0 \\
macro avg    &    92.0186 &  90.2587 &   90.1411 &    10000 \\
weighted avg &    92.3031 &  88.9900 &   89.7083 &    10000 \\
\bottomrule
\end{tabular}

    \caption{\texttt{langid.py} on Wikipedia samples of size 256}
    \label{tab:langid256}
\end{table}

\begin{table}[h]
    \centering
    \begin{tabular}{lrrrr}
\toprule
{} &  precision &  recall &  f1-score &  support \\
\midrule
ar           &      98.56 &   97.72 &     98.14 &      351 \\
de           &      64.65 &   85.91 &     73.78 &      447 \\
el           &      98.57 &   95.17 &     96.84 &      290 \\
en           &      26.52 &   88.89 &     40.85 &      369 \\
es           &      51.20 &   79.80 &     62.38 &      401 \\
fr           &      65.95 &   82.29 &     73.22 &      525 \\
he           &      99.46 &   98.40 &     98.93 &      376 \\
hi           &      99.55 &   99.55 &     99.55 &      221 \\
id           &      88.19 &   53.51 &     66.60 &      656 \\
it           &      72.24 &   68.25 &     70.19 &      652 \\
ja           &      92.34 &   91.56 &     91.95 &      474 \\
ko           &      99.69 &   97.09 &     98.37 &      652 \\
mk           &      93.42 &   77.60 &     84.78 &      366 \\
nl           &      86.16 &   61.71 &     71.91 &      585 \\
pt           &      84.43 &   65.31 &     73.65 &      490 \\
ru           &      77.71 &   90.29 &     83.53 &      278 \\
sl           &      90.05 &   63.88 &     74.74 &      623 \\
sq           &      96.94 &   63.39 &     76.65 &      549 \\
th           &     100.00 &   93.26 &     96.51 &      267 \\
tl           &      82.00 &   33.70 &     47.77 &      365 \\
vi           &      99.23 &   72.71 &     83.93 &      535 \\
zh           &      90.98 &   91.67 &     91.32 &      528 \\
accuracy     &      77.43 &   77.43 &     77.43 &        0 \\
macro avg    &      84.45 &   79.62 &     79.80 &    10000 \\
weighted avg &      84.48 &   77.43 &     78.78 &    10000 \\
\bottomrule
\end{tabular}

    \caption{\texttt{langid.py} on Wikipedia samples of size 16}
    \label{tab:langid16}
\end{table}

\begin{table}[h]
    \centering
    \begin{tabular}{lrrrr}
\toprule
{} &  precision &  recall &  f1-score &  support \\
\midrule
ar           &      98.57 &   98.01 &     98.29 &      351 \\
de           &      73.19 &   92.84 &     81.85 &      447 \\
el           &      98.91 &   93.79 &     96.28 &      290 \\
en           &      33.33 &   94.31 &     49.26 &      369 \\
es           &      65.87 &   89.53 &     75.90 &      401 \\
fr           &      74.10 &   89.90 &     81.24 &      525 \\
he           &      99.47 &   99.47 &     99.47 &      376 \\
hi           &      99.10 &   99.10 &     99.10 &      221 \\
id           &      91.18 &   66.16 &     76.68 &      656 \\
it           &      80.73 &   77.76 &     79.22 &      652 \\
ja           &      96.05 &   92.41 &     94.19 &      474 \\
ko           &      99.84 &   97.39 &     98.60 &      652 \\
mk           &      96.18 &   82.51 &     88.82 &      366 \\
nl           &      93.94 &   68.89 &     79.49 &      585 \\
pt           &      91.03 &   80.82 &     85.62 &      490 \\
ru           &      82.33 &   93.88 &     87.73 &      278 \\
sl           &      94.50 &   74.48 &     83.30 &      623 \\
sq           &      97.99 &   71.22 &     82.49 &      549 \\
th           &     100.00 &   96.63 &     98.29 &      267 \\
tl           &      87.63 &   44.66 &     59.17 &      365 \\
vi           &      99.28 &   77.38 &     86.97 &      535 \\
zh           &      92.28 &   95.08 &     93.66 &      528 \\
accuracy     &      83.71 &   83.71 &     83.71 &        0 \\
macro avg    &      88.43 &   85.28 &     85.26 &    10000 \\
weighted avg &      88.68 &   83.71 &     84.68 &    10000 \\
\bottomrule
\end{tabular}

    \caption{\texttt{langid.py} on Wikipedia samples of size 32}
    \label{tab:langid32}
\end{table}

\begin{table}[h]
    \centering
    \begin{tabular}{lrrrr}
\toprule
{} &  precision &  recall &  f1-score &  support \\
\midrule
ar           &      98.58 &   98.86 &     98.72 &      351 \\
de           &      75.91 &   93.74 &     83.88 &      447 \\
el           &      98.58 &   95.86 &     97.20 &      290 \\
en           &      36.72 &   95.93 &     53.11 &      369 \\
es           &      72.58 &   91.77 &     81.06 &      401 \\
fr           &      76.38 &   92.38 &     83.62 &      525 \\
he           &      99.21 &   99.73 &     99.47 &      376 \\
hi           &      99.55 &   99.55 &     99.55 &      221 \\
id           &      94.80 &   69.51 &     80.21 &      656 \\
it           &      83.87 &   80.52 &     82.16 &      652 \\
ja           &      96.92 &   92.83 &     94.83 &      474 \\
ko           &      99.84 &   97.85 &     98.84 &      652 \\
mk           &      96.88 &   84.97 &     90.54 &      366 \\
nl           &      95.03 &   71.97 &     81.91 &      585 \\
pt           &      93.01 &   86.94 &     89.87 &      490 \\
ru           &      85.16 &   94.96 &     89.80 &      278 \\
sl           &      96.25 &   78.33 &     86.37 &      623 \\
sq           &      98.58 &   76.14 &     85.92 &      549 \\
th           &     100.00 &   97.38 &     98.67 &      267 \\
tl           &      91.98 &   53.42 &     67.59 &      365 \\
vi           &      99.29 &   78.50 &     87.68 &      535 \\
zh           &      92.38 &   96.40 &     94.35 &      528 \\
accuracy     &      86.17 &   86.17 &     86.17 &        0 \\
macro avg    &      90.07 &   87.62 &     87.52 &    10000 \\
weighted avg &      90.36 &   86.17 &     87.01 &    10000 \\
\bottomrule
\end{tabular}

    \caption{\texttt{langid.py} on Wikipedia samples of size 64}
    \label{tab:langid64}
\end{table}

\section{Full tables : Twitter}

\subsection{SVC}

\begin{table}[h]
    \centering
    \begin{tabular}{lrrrr}
\toprule
{} &  precision &  recall &  f1-score &  support \\
\midrule
ar           &     100.00 &   97.18 &     98.57 &      497 \\
de           &      97.99 &   89.78 &     93.70 &      489 \\
el           &     100.00 &   98.17 &     99.08 &      493 \\
en           &      84.14 &   87.88 &     85.97 &      495 \\
es           &      95.73 &   82.62 &     88.69 &      489 \\
fr           &      95.48 &   90.24 &     92.79 &      492 \\
he           &     100.00 &  100.00 &    100.00 &      496 \\
hi           &     100.00 &  100.00 &    100.00 &       30 \\
id           &      89.14 &   81.07 &     84.91 &      486 \\
it           &      95.32 &   88.25 &     91.65 &      485 \\
ja           &      99.59 &   97.99 &     98.78 &      497 \\
ko           &      99.60 &   99.60 &     99.60 &      496 \\
mk           &      78.33 &   90.38 &     83.93 &       52 \\
nl           &      93.17 &   78.93 &     85.46 &      484 \\
pt           &      93.67 &   78.57 &     85.46 &      490 \\
ru           &     100.00 &   86.42 &     92.72 &      486 \\
sl           &      47.01 &   94.83 &     62.86 &       58 \\
sq           &      67.57 &   83.33 &     74.63 &       90 \\
th           &     100.00 &   98.19 &     99.09 &      498 \\
tl           &      87.80 &   84.71 &     86.23 &      340 \\
vi           &      96.29 &   92.26 &     94.23 &      478 \\
zh           &      99.79 &   95.71 &     97.70 &      489 \\
accuracy     &      90.55 &   90.55 &     90.55 &        0 \\
macro avg    &      91.67 &   90.56 &     90.55 &     8910 \\
weighted avg &      95.44 &   90.55 &     92.77 &     8910 \\
\bottomrule
\end{tabular}

    \caption{RRC on Twitter samples of size 256}
    \label{tab:tw_rrc256}
\end{table}

\begin{table}[h]
    \centering
    \begin{tabular}{lrrrr}
\toprule
{} &  precision &  recall &  f1-score &  support \\
\midrule
ar           &     100.00 &   95.77 &     97.84 &      497 \\
de           &      95.54 &   70.14 &     80.90 &      489 \\
el           &      99.79 &   94.73 &     97.19 &      493 \\
en           &      69.87 &   74.95 &     72.32 &      495 \\
es           &      85.63 &   59.71 &     70.36 &      489 \\
fr           &      86.01 &   67.48 &     75.63 &      492 \\
he           &     100.00 &   99.19 &     99.60 &      496 \\
hi           &     100.00 &  100.00 &    100.00 &       30 \\
id           &      79.31 &   61.52 &     69.29 &      486 \\
it           &      84.02 &   67.22 &     74.68 &      485 \\
ja           &      99.58 &   95.98 &     97.75 &      497 \\
ko           &      99.60 &   99.40 &     99.50 &      496 \\
mk           &      63.79 &   71.15 &     67.27 &       52 \\
nl           &      80.71 &   65.70 &     72.44 &      484 \\
pt           &      80.33 &   59.18 &     68.16 &      490 \\
ru           &      99.67 &   61.32 &     75.92 &      486 \\
sl           &      31.68 &   87.93 &     46.58 &       58 \\
sq           &      42.06 &   58.89 &     49.07 &       90 \\
th           &     100.00 &   92.17 &     95.92 &      498 \\
tl           &      77.09 &   62.35 &     68.94 &      340 \\
vi           &      91.31 &   85.77 &     88.46 &      478 \\
zh           &      99.77 &   90.39 &     94.85 &      489 \\
accuracy     &      78.20 &   78.20 &     78.20 &        0 \\
macro avg    &      84.65 &   78.07 &     79.97 &     8910 \\
weighted avg &      89.72 &   78.20 &     83.01 &     8910 \\
\bottomrule
\end{tabular}

    \caption{RRC on Twitter samples of size 16}
    \label{tab:tw_rrc16}
\end{table}

\begin{table}[h]
    \centering
    \begin{tabular}{lrrrr}
\toprule
{} &  precision &  recall &  f1-score &  support \\
\midrule
ar           &      99.38 &   96.18 &     97.75 &      497 \\
de           &      86.13 &   60.94 &     71.38 &      489 \\
el           &      99.15 &   94.32 &     96.67 &      493 \\
en           &      81.60 &   26.87 &     40.43 &      495 \\
es           &      93.20 &   39.26 &     55.25 &      489 \\
fr           &      87.54 &   52.85 &     65.91 &      492 \\
he           &     100.00 &   98.19 &     99.08 &      496 \\
hi           &     100.00 &   96.67 &     98.31 &       30 \\
id           &      48.36 &   72.63 &     58.06 &      486 \\
it           &      84.27 &   61.86 &     71.34 &      485 \\
ja           &     100.00 &   83.50 &     91.01 &      497 \\
ko           &      87.53 &   72.18 &     79.12 &      496 \\
mk           &      57.97 &   76.92 &     66.12 &       52 \\
nl           &      25.02 &   63.84 &     35.95 &      484 \\
pt           &      65.53 &   58.98 &     62.08 &      490 \\
ru           &      98.83 &   86.63 &     92.32 &      486 \\
sl           &      10.69 &   87.93 &     19.07 &       58 \\
sq           &      29.08 &   45.56 &     35.50 &       90 \\
th           &     100.00 &   92.77 &     96.25 &      498 \\
tl           &      51.83 &   54.12 &     52.95 &      340 \\
vi           &      68.53 &   74.27 &     71.29 &      478 \\
zh           &      93.47 &   76.07 &     83.88 &      489 \\
accuracy     &      70.62 &   70.62 &     70.62 &        0 \\
macro avg    &      75.82 &   71.48 &     69.99 &     8910 \\
weighted avg &      81.25 &   70.62 &     73.11 &     8910 \\
\bottomrule
\end{tabular}

    \caption{RRC on Twitter samples of size 32}
    \label{tab:tw_rrc32}
\end{table}

\begin{table}[h]
    \centering
    \begin{tabular}{lrrrr}
\toprule
{} &  precision &  recall &  f1-score &  support \\
\midrule
ar           &      98.76 &   96.38 &     97.56 &      497 \\
de           &      89.03 &   71.37 &     79.23 &      489 \\
el           &      99.37 &   95.94 &     97.63 &      493 \\
en           &      88.32 &   35.15 &     50.29 &      495 \\
es           &      94.38 &   51.53 &     66.67 &      489 \\
fr           &      92.94 &   64.23 &     75.96 &      492 \\
he           &     100.00 &   98.19 &     99.08 &      496 \\
hi           &     100.00 &  100.00 &    100.00 &       30 \\
id           &      55.18 &   80.04 &     65.32 &      486 \\
it           &      87.43 &   68.87 &     77.05 &      485 \\
ja           &     100.00 &   86.32 &     92.66 &      497 \\
ko           &      90.17 &   73.99 &     81.28 &      496 \\
mk           &      66.13 &   78.85 &     71.93 &       52 \\
nl           &      31.74 &   64.67 &     42.59 &      484 \\
pt           &      69.70 &   65.71 &     67.65 &      490 \\
ru           &      98.44 &   91.15 &     94.66 &      486 \\
sl           &      12.79 &   96.55 &     22.58 &       58 \\
sq           &      40.13 &   70.00 &     51.01 &       90 \\
th           &     100.00 &   94.78 &     97.32 &      498 \\
tl           &      58.06 &   63.53 &     60.67 &      340 \\
vi           &      73.65 &   77.20 &     75.38 &      478 \\
zh           &      94.67 &   79.96 &     86.70 &      489 \\
accuracy     &      75.93 &   75.93 &     75.93 &        0 \\
macro avg    &      79.13 &   77.47 &     75.15 &     8910 \\
weighted avg &      84.14 &   75.93 &     77.98 &     8910 \\
\bottomrule
\end{tabular}

    \caption{RRC on Twitter samples of size 64}
    \label{tab:tw_rrc64}
\end{table}

\subsection{SVC}
\begin{table}[h]
    \centering
    \begin{tabular}{lrrrr}
\toprule
{} &  precision &  recall &  f1-score &  support \\
\midrule
ar           &      98.98 &   97.18 &     98.07 &      497 \\
de           &      91.80 &   80.16 &     85.59 &      489 \\
el           &      99.38 &   96.96 &     98.15 &      493 \\
en           &      91.15 &   35.35 &     50.95 &      495 \\
es           &      97.95 &   58.49 &     73.24 &      489 \\
fr           &      94.85 &   71.14 &     81.30 &      492 \\
he           &     100.00 &   98.59 &     99.29 &      496 \\
hi           &     100.00 &  100.00 &    100.00 &       30 \\
id           &      59.80 &   86.63 &     70.76 &      486 \\
it           &      88.34 &   78.14 &     82.93 &      485 \\
ja           &     100.00 &   88.73 &     94.03 &      497 \\
ko           &      72.88 &   85.08 &     78.51 &      496 \\
mk           &      70.97 &   84.62 &     77.19 &       52 \\
nl           &      53.86 &   59.09 &     56.35 &      484 \\
pt           &      71.15 &   67.96 &     69.52 &      490 \\
ru           &      98.92 &   94.24 &     96.52 &      486 \\
sl           &      12.50 &   96.55 &     22.13 &       58 \\
sq           &      36.31 &   72.22 &     48.33 &       90 \\
th           &     100.00 &   95.38 &     97.64 &      498 \\
tl           &      59.85 &   69.71 &     64.40 &      340 \\
vi           &      71.68 &   77.82 &     74.62 &      478 \\
zh           &      93.53 &   85.69 &     89.43 &      489 \\
accuracy     &      79.58 &   79.58 &     79.58 &        0 \\
macro avg    &      80.18 &   80.90 &     77.68 &     8910 \\
weighted avg &      85.29 &   79.58 &     80.85 &     8910 \\
\bottomrule
\end{tabular}

    \caption{SVC on Twitter samples of size 256}
    \label{tab:tw_svc256}
\end{table}

\begin{table}[h]
    \centering
    \begin{tabular}{lrrrr}
\toprule
{} &  precision &  recall &  f1-score &  support \\
\midrule
ar           &      99.58 &   95.17 &     97.33 &      497 \\
de           &      79.92 &   43.15 &     56.04 &      489 \\
el           &      99.12 &   90.87 &     94.81 &      493 \\
en           &      76.85 &   16.77 &     27.53 &      495 \\
es           &      77.25 &   33.33 &     46.57 &      489 \\
fr           &      82.21 &   42.28 &     55.84 &      492 \\
he           &     100.00 &   96.98 &     98.46 &      496 \\
hi           &     100.00 &   96.67 &     98.31 &       30 \\
id           &      41.35 &   61.93 &     49.59 &      486 \\
it           &      76.09 &   43.30 &     55.19 &      485 \\
ja           &      99.51 &   81.69 &     89.72 &      497 \\
ko           &      87.86 &   67.14 &     76.11 &      496 \\
mk           &      44.12 &   57.69 &     50.00 &       52 \\
nl           &      18.10 &   60.12 &     27.82 &      484 \\
pt           &      58.09 &   41.02 &     48.09 &      490 \\
ru           &      97.12 &   83.13 &     89.58 &      486 \\
sl           &       9.52 &   86.21 &     17.15 &       58 \\
sq           &      25.83 &   34.44 &     29.52 &       90 \\
th           &     100.00 &   88.35 &     93.82 &      498 \\
tl           &      43.72 &   47.06 &     45.33 &      340 \\
vi           &      57.63 &   70.29 &     63.34 &      478 \\
zh           &      92.51 &   70.76 &     80.19 &      489 \\
accuracy     &      63.24 &   63.24 &     63.24 &        0 \\
macro avg    &      71.20 &   64.02 &     63.20 &     8910 \\
weighted avg &      76.70 &   63.24 &     66.20 &     8910 \\
\bottomrule
\end{tabular}

    \caption{SVC on Twitter samples of size 16}
    \label{tab:tw_svc16}
\end{table}

\begin{table}[h]
    \centering
    \begin{tabular}{lrrrr}
\toprule
{} &  precision &  recall &  f1-score &  support \\
\midrule
ar           &      99.38 &   96.18 &     97.75 &      497 \\
de           &      86.13 &   60.94 &     71.38 &      489 \\
el           &      99.15 &   94.32 &     96.67 &      493 \\
en           &      81.60 &   26.87 &     40.43 &      495 \\
es           &      93.20 &   39.26 &     55.25 &      489 \\
fr           &      87.54 &   52.85 &     65.91 &      492 \\
he           &     100.00 &   98.19 &     99.08 &      496 \\
hi           &     100.00 &   96.67 &     98.31 &       30 \\
id           &      48.36 &   72.63 &     58.06 &      486 \\
it           &      84.27 &   61.86 &     71.34 &      485 \\
ja           &     100.00 &   83.50 &     91.01 &      497 \\
ko           &      87.53 &   72.18 &     79.12 &      496 \\
mk           &      57.97 &   76.92 &     66.12 &       52 \\
nl           &      25.02 &   63.84 &     35.95 &      484 \\
pt           &      65.53 &   58.98 &     62.08 &      490 \\
ru           &      98.83 &   86.63 &     92.32 &      486 \\
sl           &      10.69 &   87.93 &     19.07 &       58 \\
sq           &      29.08 &   45.56 &     35.50 &       90 \\
th           &     100.00 &   92.77 &     96.25 &      498 \\
tl           &      51.83 &   54.12 &     52.95 &      340 \\
vi           &      68.53 &   74.27 &     71.29 &      478 \\
zh           &      93.47 &   76.07 &     83.88 &      489 \\
accuracy     &      70.62 &   70.62 &     70.62 &        0 \\
macro avg    &      75.82 &   71.48 &     69.99 &     8910 \\
weighted avg &      81.25 &   70.62 &     73.11 &     8910 \\
\bottomrule
\end{tabular}

    \caption{SVC on Twitter samples of size 32}
    \label{tab:tw_svc32}
\end{table}

\begin{table}[h]
    \centering
    \begin{tabular}{lrrrr}
\toprule
{} &  precision &  recall &  f1-score &  support \\
\midrule
ar           &      98.76 &   96.38 &     97.56 &      497 \\
de           &      89.03 &   71.37 &     79.23 &      489 \\
el           &      99.37 &   95.94 &     97.63 &      493 \\
en           &      88.32 &   35.15 &     50.29 &      495 \\
es           &      94.38 &   51.53 &     66.67 &      489 \\
fr           &      92.94 &   64.23 &     75.96 &      492 \\
he           &     100.00 &   98.19 &     99.08 &      496 \\
hi           &     100.00 &  100.00 &    100.00 &       30 \\
id           &      55.18 &   80.04 &     65.32 &      486 \\
it           &      87.43 &   68.87 &     77.05 &      485 \\
ja           &     100.00 &   86.32 &     92.66 &      497 \\
ko           &      90.17 &   73.99 &     81.28 &      496 \\
mk           &      66.13 &   78.85 &     71.93 &       52 \\
nl           &      31.74 &   64.67 &     42.59 &      484 \\
pt           &      69.70 &   65.71 &     67.65 &      490 \\
ru           &      98.44 &   91.15 &     94.66 &      486 \\
sl           &      12.79 &   96.55 &     22.58 &       58 \\
sq           &      40.13 &   70.00 &     51.01 &       90 \\
th           &     100.00 &   94.78 &     97.32 &      498 \\
tl           &      58.06 &   63.53 &     60.67 &      340 \\
vi           &      73.65 &   77.20 &     75.38 &      478 \\
zh           &      94.67 &   79.96 &     86.70 &      489 \\
accuracy     &      75.93 &   75.93 &     75.93 &        0 \\
macro avg    &      79.13 &   77.47 &     75.15 &     8910 \\
weighted avg &      84.14 &   75.93 &     77.98 &     8910 \\
\bottomrule
\end{tabular}

    \caption{SVC on Twitter samples of size 64}
    \label{tab:tw_svc64}
\end{table}

\subsection{RRC + Fasttext}
\begin{table}[h]
    \centering
    \begin{tabular}{lrrrr}
\toprule
{} &  precision &  recall &  f1-score &  support \\
\midrule
ar           &      98.98 &   97.18 &     98.07 &      497 \\
de           &      92.92 &   91.21 &     92.05 &      489 \\
el           &      99.59 &   98.17 &     98.88 &      493 \\
en           &      71.43 &   91.92 &     80.39 &      495 \\
es           &      94.53 &   84.87 &     89.44 &      489 \\
fr           &      91.35 &   92.28 &     91.81 &      492 \\
he           &     100.00 &  100.00 &    100.00 &      496 \\
hi           &     100.00 &  100.00 &    100.00 &       30 \\
id           &      88.82 &   83.33 &     85.99 &      486 \\
it           &      91.36 &   91.55 &     91.45 &      485 \\
ja           &      97.43 &   98.99 &     98.20 &      497 \\
ko           &      99.00 &   99.60 &     99.30 &      496 \\
mk           &      78.69 &   92.31 &     84.96 &       52 \\
nl           &      92.73 &   84.30 &     88.31 &      484 \\
pt           &      93.27 &   82.04 &     87.30 &      490 \\
ru           &      99.16 &   96.91 &     98.02 &      486 \\
sl           &      45.45 &   94.83 &     61.45 &       58 \\
sq           &      66.37 &   83.33 &     73.89 &       90 \\
th           &      99.80 &   98.19 &     98.99 &      498 \\
tl           &      86.49 &   84.71 &     85.59 &      340 \\
vi           &      96.10 &   92.68 &     94.36 &      478 \\
zh           &      98.35 &   97.75 &     98.05 &      489 \\
accuracy     &      92.65 &   92.65 &     92.65 &        0 \\
macro avg    &      90.08 &   92.55 &     90.75 &     8910 \\
weighted avg &      93.43 &   92.65 &     92.84 &     8910 \\
\bottomrule
\end{tabular}

    \caption{RRC + FastText on Twitter samples of size 256}
    \label{tab:tw_hybrid256}
\end{table}

\begin{table}[h]
    \centering
    %

    \caption{RRC + FastText on Twitter samples of size 16}
    \label{tab:tw_hybrid16}
\end{table}

\begin{table}[h]
    \centering
    %

    \caption{RRC + FastText on Twitter samples of size 32}
    \label{tab:tw_hybrid32}
\end{table}

\begin{table}[h]
    \centering
    %

    \caption{RRC + FastText on Twitter samples of size 64}
    \label{tab:tw_hybrid64}
\end{table}

\subsection{FastText}
\begin{table}[h]
    \centering
    %

    \caption{FastText on Twitter samples of size 256}
    \label{tab:tw_fasttext256}
\end{table}

\begin{table}[h]
    \centering
    %

    \caption{FastText on Twitter samples of size 16}
    \label{tab:tw_fasttext16}
\end{table}

\begin{table}[h]
    \centering
    %

    \caption{FastText on Twitter samples of size 32}
    \label{tab:tw_fasttext32}
\end{table}

\begin{table}[h]
    \centering
    %

    \caption{FastText on Twitter samples of size 64}
    \label{tab:tw_fasttext64}
\end{table}

\subsection{\texttt{langid.py}}
\begin{table}[h]
    \centering
    %

    \caption{\texttt{langid.py} on Twitter samples of size 256}
    \label{tab:tw_langid256}
\end{table}

\begin{table}[h]
    \centering
    %

    \caption{\texttt{langid.py} on Twitter samples of size 16}
    \label{tab:tw_langid16}
\end{table}

\begin{table}[h]
    \centering
    %

    \caption{\texttt{langid.py} on Twitter samples of size 32}
    \label{tab:tw_langid32}
\end{table}

\begin{table}[h]
    \centering
    %

    \caption{\texttt{langid.py} on Twitter samples of size 64}
    \label{tab:tw_langid64}
\end{table}

\end{document}